\documentclass[12pt]{article}
\usepackage[sort&compress,numbers]{natbib}
\usepackage{amsmath}
\usepackage{amsfonts}
\usepackage{graphicx}
\usepackage{hyperref}
\graphicspath{{./figures/}}
\usepackage{epsfig}
\usepackage{color}
\usepackage{xr}
\usepackage{bm}
\usepackage{mathtools}
\mathtoolsset{showonlyrefs}
\usepackage[normalem]{ulem}
\usepackage{dsfont}
\usepackage{authblk}
\usepackage[margin=1in,showframe=false]{geometry}

\newcommand{\REVISION}[1]{{\color{black}{#1}}}

\title{A Survey of Constrained Gaussian Process Regression: \\ Approaches and Implementation Challenges}

\author[1]{L.~Swiler}
\author[1]{M.~Gulian}
\author[2]{A.~Frankel}
\author[2]{C.~Safta}
\author[1]{J.~Jakeman}
\affil[1]{\small Center for Computing Research, Sandia National Laboratories, Albuquerque, NM, 87123}
\affil[2]{\small Computational Science and Analysis, Sandia National Laboratories, Livermore, CA, 94550}
\affil[ ]{\small \texttt{\{lpswile,mgulian,alfrank,csafta,jdjakem\}@sandia.gov}}
 
\begin{document}

\maketitle
%\tableofcontents

\begin{abstract}
Gaussian process regression is a popular Bayesian framework for surrogate modeling of expensive data sources. As part of a broader effort in scientific machine learning, many recent works have incorporated physical constraints or other a priori information within Gaussian process regression to supplement limited data and regularize the behavior of the model.
We provide an overview and survey of several classes of Gaussian process constraints, including positivity or bound constraints, monotonicity and convexity constraints, differential equation constraints provided by linear PDEs, and boundary condition constraints. We compare the strategies behind each approach as well as the differences in implementation, concluding with a discussion of the computational challenges introduced by constraints.   

\end{abstract}

\section{Introduction}
\label{sec:intro}

There has been a tremendous surge in the development and application of machine learning models in recent years due to their flexibility and capability to represent trends in complex systems \citep{hastie2016elements}.
The parameters of a machine learning model can often be calibrated, with sufficient data, to give high fidelity representations of the underlying process %being approximated
\citep{raissi2017,jones2018machine,frankel2019predicting,frankel2019prediction}.
It is now feasible to construct deep learning models over datasets of tens of thousands to millions of data points with modern computational resources  \citep{dean2012large}. In many scientific applications, however, there may not be large amount\REVISION{s} of data available for training. Unlike data from internet or text searches, computational and physical experiments are typically extremely expensive. Moreover, even if ample data exists, the machine learning model may yield behaviors that are inconsistent with what is expected physically when queried in an extrapolatory regime.

To aid and improve the process of building machine learning models for scientific applications, it is desirable to have a framework that allows the incorporation of physical principles and other a priori information to supplement the limited data and regularize the behavior of the model.  Such a framework is often referred to as ``physics-constrained'' machine learning within the scientific computing community \citep{raissi2018, pan2018data, lusch2018deep, brunton2016discovering, LeeCarlberg, ling2016machine,jones2018machine}.
\citet{karpatne2017theory} provide a taxonomy for theory-guided data science, 
with the goal of incorporating scientific
consistency in the learning of generalizable models.  The information used to constrain models can be simple, such as known range or positivity constraints, shape constraints, or monotonicity constraints that the machine learning model must satisfy.  The constraints can also be more complex; for example, they can encode knowledge of the underlying data-generating process in the form of a partial differential equation. 
%To highlight the interest in ``physics-informed'' machine learning, there is a curated bibliography maintained by \citet{constantine_website} that has over 200 references to papers involving scientific machine learning.
Several recent conferences highlight the interest in ``physics-informed'' machine learning~\citep{conference1,conference2,conference3,conference4,conference5}. 

Much of the existing research in physics-informed machine learning has focused on incorporating constraints in neural networks \citep{ling2016machine,jones2018machine}, often through the use of objective/loss functions which penalize constraint violation \citep{raissi2018deep,raissi2019physics,magiera2019constraint,cyr2019robust,mao2020physics}. Other works have focused on incorporating prior knowledge using Bayesian inference that expresses the data-generating process as dependent on a set of parameters, the initial distribution of which is determined by the available information, e.g., functional constraints \citep{wangBerger, jidling2017linearly}. Unlike deterministic learning approaches, the predictions made using approximations trained with Bayesian inference are accompanied with probabilistic estimates of uncertainty/error.

Within the Bayesian regression framework, Gaussian processes (GPs) are popular for constructing ``surrogates'' or ``emulators'' of data sources that are very expensive to query.  The use of GPs in a 
regression framework to predict a set of function values is called Gaussian Process Regression (GPR). 
An accurate GPR can often be constructed using only a relatively small number of training data (e.g. tens to hundreds), which consists of pairs of input 
parameters and corresponding response values.  Once constructed, the GPR can be thought of as a machine-learned metamodel and used to provide 
fast, cheap function evaluations for the purposes of prediction, sensitivity 
analysis, uncertainty quantification, calibration, and optimization. 
GP regression models are constructed with data obtained from computational simulation~\citep{gramacy2020surrogates} or field data; in geostatistics, the process of applying Gaussian processes to field data has been used for decades and is frequently referred to as \emph{kriging} \citep{chiles2018fifty}. 
%Seminal works on the use of Gaussian processes as surrogate models 
%for computational science and engineering simulations include ~\cite{sacks},~\cite{santner},~\cite{simpson}.

In this survey we focus on the use of constrained GPRs
%as surrogates for simulations of physical systems that are governed, for example, by partial differential equations. 
%The approaches discussed in this paper represent a broad set of methods used to construct Gaussian process models
that honor or incorporate a wide variety of physical constraints~\citep{riihimaki, daveiga, jensen, solak, raissi2017, yang, lopez2018, bachoc2019}.
Specifically,
%we provide an overview of Gaussian process constraints including: positivity or bounds constraints on the machine learning model itself and/or its derivatives; linear equality or inequality constraints, conservation laws; and rotational symmetries or invariances. 
we focus on the following topics, after a short review of Gaussian process regression in Section \ref{gpr}. 
Section \ref{overview_of_constraints} presents an overview and a classification of constraints according to how the constraint is enforced during the construction of a GP. 
Section \ref{sec:bound_constraints} discusses bound constraints, in which the GP prediction may be required to be positive, for example, or the prediction may be required to fall between upper and lower bounds.
Section \ref{sec:monotonicity_constraints} discusses monotonicity and related convexity constraints.
Constraints may also be more 
tightly integrated with the underlying physics:  the GP can be constrained to satisfy 
linear operator constraints which represent physical laws expressed as partial different equations (PDE). 
This is discussed in Section~\ref{sec:pde_constraints}.  
Section \ref{sec:boundary_constraints} discusses intrinsic boundary condition constraints.
We review several different approaches for enforcing each of these constraint types. 
Finally, Section \ref{sec:computation_considerations} is a compendium of computational details for implementing the constraints of Sections \ref{sec:bound_constraints} -- \ref{sec:boundary_constraints}, together with a summary of computational strategies for improving GPR and brief commentary about the challenges of applying these strategies for the constrained GPs considered here.

The taxonomy we present is formulated to enable practitioners to easily query this overview for information on the specific constraint(s) they may be interested in. 
For approaches that enforce different constraints but have significant overlap in methodology, references are made between sections to the prerequisite subsection where the technical basis of an approach is first discussed in detail. This is done, for example, when discussing spline-based approaches which are used for both bound constraints in Section \ref{sec:bound_constraints} and monotonicity constraints in Section \ref{sec:monotonicity_constraints}.

Not all physical constraints can be neatly divided into the categories that we focus on in Sections \ref{sec:bound_constraints} -- \ref{sec:boundary_constraints}. For example, with a view toward computer vision, \citet{salzmann2010implicitly} considered GPR for pose estimation under rigid (constant angle and length) and non-rigid (constant length) constraints between points. They proved that linear equality constraints of the form $A\mathbf{y} = \mathbf{b}$, if satisfied by all the data vectors $\mathbf{y}$, are satisfied by the posterior mean predictor of a GP. Then, at the cost of squaring the input dimension, they translated quadratic length constraints into such linear constraints for pairwise products of the input variables. 
In another example, \citet{frankel2019tensor} applied GPR to predict the behavior of hyperelastic materials, in which the stress-stretch constitutive relation naturally exhibits rotational invariance. The rotational invariance was enforced by deriving a finite expansion of the Cauchy stress tensor in powers of the Finger tensor that satisfies the rotational invariance by virtue of its structure, and GPR was performed for the coefficients of the expansion.  
We mention these examples to illustrate that physical constraints are varied, and in some cases the method to enforce them can depend highly on the specific nature of the constraint.

Even within the selected categories represented by Sections \ref{sec:bound_constraints} -- \ref{sec:boundary_constraints}, the literature on constrained Gaussian processes is extensive and expanding rapidly. Consequently, we cannot provide a complete survey of every instance of constrained GPR.  
Rather, we strive to discuss main areas of research within the field. The goal is to aid readers in selecting methods appropriate for their applications and enable further exploration of the literature. We present selected implementation details and numerical examples, giving references to the original works for further details. Many of the authors of these works have developed codebases and released them publicly. Finally, we remark that we have adopted consistent notation (established in Section \ref{gpr}) for GPR that does not always follow the notation of the original works exactly.

\section{Gaussian Process Regression}\label{gpr}
\label{sec:gpr}
This section provides an overview of unconstrained Gaussian process regression. 
As mentioned previously, Gaussian process models, or simply Gaussian processes, are popular because they can be used in a regression framework to approximate 
complicated nonlinear functions with probabilistic estimates of the uncertainty. Seminal work discussing the use of GPs as surrogate models for computational science and engineering applications include the papers of \citet{sacks} and \citet{santner} and the book by \citet{rasmussen}.

A Gaussian process can be viewed as a distribution over a set of functions.
A random draw or sample $f$ from a GP is a realization from the set of admissible functions. 
Specifically, a Gaussian process is a collection of random variables
$\{f(\mathbf{x}) \ | \ \mathbf{x} \in \REVISION{\Omega \subset \mathbb{R}^d} \}$ for which, given any finite set of $N$ inputs $X = \{\mathbf{x}_1, \mathbf{x}_2, ..., \mathbf{x}_N\}$, $\mathbf{x}_i \in \REVISION{\Omega}$, the collection $f(\mathbf{x}_1),f(\mathbf{x}_2),...,f(\mathbf{x}_N)$  
has a joint multivariate Gaussian distribution.  
A GP is completely defined by its mean and covariance functions which generate the mean vectors and covariances matrices of these finite-dimensional multivariate normals. Assumptions such as smoothness of $f$, stationarity, and sparsity are used to construct the mean and covariance of the GP prior and then Bayes' rule is used to constrain the prior on observational/simulation data.

The prediction 
$\mathbf{f}=[f(\bold{x}_1),f(\mathbf{x}_2),...f(\mathbf{x}_N)]^\top$ of a Gaussian process with mean function 
$m(\mathbf{x})$
and a covariance function $k(\mathbf{x},\mathbf{x}')$ is a random variable such that
\begin{equation}
p(\bold{f}|X) = \mathcal{N}(\mathbf{f}; m(X), k(X,X)),
\label{eq:multivar_norm}
\end{equation}
where $m(X)$ denotes the vector $[m(\mathbf{x}_1), ..., m(\mathbf{x}_N)]^\top$ and $k(X,X)$ denotes the matrix with entries $[k(\mathbf{x}_i,\mathbf{x}_j)]_{1 \le i,j \le N}$. 
The multivariate normal probability \REVISION{density function}
$\mathcal{N}(\mathbf{f}; \mathbf{m}, K)$ with mean vector $\mathbf{m}$ and covariance matrix $K$ has the form
\begin{equation}\label{standard_mvn}
\mathcal{N}(\mathbf{f}; \mathbf{m}, K) = \frac{1}{(2\pi)^{N/2}|{K^{1/2}|}}\exp\left({-\frac{1}{2}(\mathbf{f}-\mathbf{m})^\top K^{-1}(\mathbf{f}-\mathbf{m})}\right).
\end{equation}
The covariance kernel function $k$ of a Gaussian process must be symmetric and positive semidefinite. Denoting the individual $d$ components of the vector $\mathbf{x}_{i}$ as ${x_i}^\ell$, where $\ell=1,...,d$, the squared exponential kernel
\begin{equation} 
k(\bold{x}_{i},\mathbf{x}_{j}) = \eta^2\exp\left[-\frac{1}{2}\sum_{\ell=1}^{d}
\left(\frac{{x_i}^{\ell}-{x_j}^{\ell}}{\rho_\ell}\right)^2\right], \quad \REVISION{\eta, \rho_1, ..., \rho_d \in \mathbb{R},}
\label{eq:gpkernel}
\end{equation}
is popular, but many other covariance kernels are available. The choice of a covariance kernel can have profound impact on the GP predictions \citep{kernel_cookbook,rasmussen}, and several approaches to constraining GPs that we survey rely on construction
of a covariance kernel specific to the constraint.

The \REVISION{density} \eqref{eq:multivar_norm}, determined by covariance kernel $k$ and the mean $m$, is referred to as a \emph{prior} for the GP. If the error or noise relating the actual observations $\mathbf{y} = [y(\bold{x}_1),y(\mathbf{x}_2),...,y(\mathbf{x}_N)]^\top$ collected at the set of inputs $X=\{\mathbf{x}_i\}_{i=1}^N$ to the GP prediction $\mathbf{f}$ is assumed to be Gaussian, then the probability of observing data 
$\mathbf{y}$ given the GP prior is given by
\begin{equation}\label{eq:gauss_like}
p(\bold{y}|X,\bold{f}) = \mathcal{N}(\REVISION{\bold{y}}; \mathbf{f},\sigma^2 I_N), \quad \REVISION{\sigma \in \mathbb{R}}.
\end{equation}
Here, $I_N$ denotes the $N \times N$ identity matrix. The \REVISION{density} $p(\bold{y}|X,\bold{f})$ is referred to the \emph{likelihood} of the GP, and the Gaussian likelihood \eqref{eq:gauss_like} is by far the most common. As discussed in Section~\ref{sec:likelihood_formulations}, specific non-Gaussian likelihood functions can be used to enforce certain types of constraints.

The parameters in the covariance kernel function of a GP are referred to as \emph{hyperparameters} of the GP. We denote them by $\bm{\theta}$.
For the squared exponential kernel \eqref{eq:gpkernel}, the aggregate vector of hyperparameters is  
$\bm{\theta}=[\eta, \rho_1, ..., \rho_d, \sigma]$, where we have included the likelihood/noise parameter $\sigma$ from \eqref{eq:gauss_like} as a hyperparameter. In general, finding the best hyperparameters to fit the data is an important step of GPR known as \emph{training}. From now on, we explicitly denote the dependence on $\bm{\theta}$ of the likelihood $p(\bold{y}|X,\bold{f})$ in \eqref{eq:gauss_like} and the prior $p(\bold{f}|X)$ in \eqref{eq:multivar_norm}, writing these as $p(\bold{y}|X,\bold{f}, \bm{\theta})$ and $p(\bold{f}|X, \bm{\theta})$, respectively. 
The marginal likelihood is given by 
\begin{equation}\label{eq:marginal-likelihood}
p(\bold{y}|X,\bm{\theta})=\int{p(\bold{y}|X,\bold{f},\bm{\theta})p(\bold{f}|X,\bm{\theta}) d\bold{f}}
\end{equation} 
and the log-marginal-likelihood for a GP with a zero-mean prior ($m \equiv 0$) can be written \citep{murphy2012machine,rasmussen} as
\begin{equation}
\log{p(\bold{y}|X,\bm{\theta)}}=-\frac{1}{2}\bold{y^\top}\big(K(X,X)
+\sigma^2 I_N\big)^{-1}\bold{y}-\frac{1}{2}\log\left|K(X,X)
+\sigma^2 I_N\right|-\frac{N}{2}\log{2\pi}
\label{eq:log_like}
\end{equation}
Formula \eqref{eq:log_like}, derived from \eqref{eq:marginal-likelihood}, \eqref{eq:multivar_norm} and \eqref{standard_mvn}, is a function of the hyperparameters $\bm{\theta}$ present in the kernel $k$, which can be optimized to give the most likely values of the hyperparameters given data. This is known as maximum likelihood estimation (MLE) of the hyperparameters. 

Once the hyperparameters of the GPR have been chosen, the \emph{posterior} of the GP is given by 
Bayes' rule,
\begin{equation}
\label{eq:bayes}
p(\bold{f} | X, \bold{y}, \bm{\theta})
=
\frac{p(\bold{f} | X,\bm{\theta}) p(\bold{y} | X, \bold{f}, \bm{\theta}) }
{p(\bold{y} | X, \bm{\theta})}.
\end{equation}
Given the prior $p(\bold{f} | X,\bm{\theta})$ \eqref{eq:multivar_norm} and the Gaussian likelihood $p(\bold{y} | X, \bold{f}, \bm{\theta})$ \eqref{eq:gauss_like}, the prediction $f^*$ of a GPR at a new point $\bold{x^*}$ can be calculated~\citep{rasmussen} as   
\REVISION{
\begin{equation}
{p({f^*}  |  \bold{y},X,\bold{x}^*,\bm{\theta})} =
\mathcal{N} (\widehat{m}(\bold{x}^*), \hat{v}(\bold{x}^*)),
\label{eq:gpreg}
\end{equation}
where
\begin{align}
\begin{split}\label{eq:post_mean_var}
\widehat{m}(\bold{x}^*) &= k(\bold{x}^*,X)(K(X,X)+\sigma^2 I_N)^{-1}\bold{y},\\
\hat{v}(\bold{x}^*) & = k(\bold{x}^*,\bold{x}^*)-k(\bold{x}^*,X)(K(X,X)+\sigma^2 I_N)^{-1}\left[k(\bold{x}^*,X)\right]^\top.
\end{split}
\end{align}
Note that the mean $\widehat{m}(\bold{x}^*)$ of this Gaussian posterior is the mean estimate $\mathbb{E}[f(\mathbf{x}^*)]$ of the predicted function value $f^*$ at $\bold{x^*}$, and the variance $\hat{v}(\bold{x}^*)$ is the estimated prediction variance of the same quantity.}

We now preface some of the computational issues of inference in GPR that will be important for the constrained case. Firstly, when the GP likelihood is Gaussian, the posterior \eqref{eq:gpreg} is also Gaussian, thus it can be computed exactly and sampling is from the posterior is simple. This is generally not the case when the likelihood is not Gaussian. The same issue arises if the \REVISION{density} \eqref{eq:gpreg} is directly replaced by a non-Gaussian \REVISION{density} in the course of enforcing constraints -- by truncation, for example. Next, inversion of $(K(X,X)+\sigma^2 I_N)$, which scales as $N^3$, is an omnipresent issue for inference. This poor scaling to large data is compounded by the fact that increased data tends to rapidly increase the condition number of $K(X,X)$ (see Section \ref{sec:subset_of_data}). Finally, optimizing the hyperparameters of the GP involves the nonconvex  objective function~\eqref{eq:log_like}; both this function and its derivatives are potentially costly and unstable to compute for the reasons just mentioned. These issues arise for conventional GPR, but throughout sections \ref{sec:bound_constraints} -- \ref{sec:boundary_constraints} we shall see that constraining the GP can make them more severe. Therefore, we review potential strategies for dealing with them in Section \ref{sec:computation_considerations}.

\section{Strategies for Constraints}\label{overview_of_constraints}
There are many ways to constrain a Gaussian process model.
The difficulty with applying constraints to a GP is that a constraint typically calls for a condition to hold \emph{globally} -- that is, for
\emph{all} points $x$ in a \REVISION{continuous domain} -- for \REVISION{\emph{all}} realizations or predictions of the process. \emph{A priori}, this amounts to an infinite set of point constraints for an infinite dimensional sample space of functions. This raises a numerical feasibility issue, which each method circumvents in some way. Some methods relax the global constraints to constraints at a finite set of ``virtual'' points; others transform the output of the GP to guarantee the predictions satisfy constraints, or construct a sample space of predictions in which every realization satisfies the constraints. 
This distinction between should be kept in mind when surveying constrained GPs. For example, the methods in Sections \ref{sec:transform_output}, \ref{sec:splines}, and \ref{transformed_covariance} enforce constraints globally. The methods in Sections \ref{sec:daveiga} and \ref{pde_constraints} enforce the constraint at scattered auxiliary data points, be this a result of introducing virtual data points for constraints, incomplete knowledge, or spatial variability.

Strategies for enforcing constraints are apparent from the review of GPR in Section \ref{gpr}, which covers posterior prediction for $\mathbf{f}$, the likelihood function for observations $\mathbf{y}$, the kernel prior $K$, and the data involved in GPR.   
Some methods, such as the warping method of Section \ref{sec:transform_output}, simply 
apply a transformation to the output $\bold{f}$ of GPR, so the transformed output satisfies the constraint. 
This transformation is essentially independent of the other components of GPR.
One can instead introduce the constraints at the prediction of $\bold{f}$, replacing the \REVISION{density} \eqref{eq:gpreg}, by augmenting the data with a discrete set of virtual points in the domain and predicting $\bold{f}$ from the GP given the data \emph{and} knowledge that the constraint holds at the virtual points. An example of this is in Section \ref{sec:daveiga}. 
Next, the likelihood $p(\bold{y}|X,\bold{f})$ provides another opportunity to enforce constraints. One can replace the Gaussian likelihood  \eqref{eq:gauss_like} by a likelihood function such that constraints are satisfied by $\bold{y}$ regardless of the output $\bold{f}$.
\REVISION{Hyperparameter optimization provides yet another opportunity, in which maximization of the marginal-log-likelihood \eqref{eq:log_like} is augmented with constraints on the posterior predictions of the GP, as in Section \ref{sec:nonneg}.}

A different strategy is to design a covariance kernel for the prior \eqref{eq:multivar_norm} of the Gaussian process which enforces the constraint.
Several of the methods discussed in this survey involve regression with an appropriate joint GP, defined by the constraint, which uses a ``four block'' covariance kernel incorporating the constraint in some of the blocks.
This is the strategy used for the linear PDE constraints in Section \ref{pde_constraints}.
Such methods are based on derivations of linear transformations of GPs. 
These types of kernels can combined with other strategies for constraints, such as for the monotonicity constraints of Section \ref{constrained_likelihood_with_derivative_information} which use a four block covariance kernel (for $\mathbf{f}$ and $\mathbf{f}'$) within a likelihood approach.

Considering Gaussian processes as distributions over functions, another strategy is to consider a function space defined by a certain representation such that a global constraint can be translated into a finite set of constraints, e.g. on the coefficients of a spline expansion in Sections \ref{sec:splines} and \ref{sec:splines_monotonic}. Or a representation can be sought such that every element of the sample space satisfies the constraint before the Gaussian process (the distribution) is even introduced. The latter approach is taken in Sections \ref{transformed_covariance} and \ref{sec:boundary_constraints}; in these cases, this strategy amounts to deriving a specific kernel function related to the representation. 

Finally, data provides an opportunity to constrain Gaussian processes implicitly. Some approaches involve proving that, if the data fed into a GP through the posterior formula \eqref{eq:bayes} satisfies the constraint, then the GP predictions satisfy the constraint -- either exactly, as for linear and quadratic equality constraints of \citet{salzmann2010implicitly}, or within a certain error, as in linear PDE constraints discussed in Section \ref{sec:empirical}. These results consider properties of GPs that form the basis of such algorithms.

\REVISION{We note that some of the methods we cover may result in a posterior distribution which is no longer Gaussian, unlike the standard GPR posterior \eqref{eq:gpreg}. Thus, such ``constrained Gaussian processes'' no longer meet the definition of a ``Gaussian process'', rendering the former term a misnomer in this strict sense. Nevertheless, we refer to any method that uses the basic steps of GPR described in Section \ref{sec:gpr} as a starting point for a constrained regression algorithm, as providing a ``constrained Gaussian process''.}

\section{Bound Constraints}\label{sec:bound_constraints}
Bound constraints of the form $a \le f(\mathbf{x}) \le b$ over some region of interest arise naturally in many applications. For example, regression over chemical concentration data should enforce that predicted values lie between $0$ and $1$ \citep{chemconcs}. Bound constraints also include, as a special case, nonnegativity constraints $f \ge 0$ ($a = 0, b = \infty$). In this section we present three approaches for enforcing bound constraints.

\subsection{Transformed Output and Likelihood}
The most direct way to impose bound constraints on a Gaussian process involves modifying the output of the regression. One way to do this is to transform the output $f$ of the GP using a ``warping'' function which satisfies the bounds. The second way is to replace the Gaussian likelihood \eqref{eq:gauss_like} by a non-Gaussian likelihood that satisfies the bounds, which is then used to obtain a posterior formula for predicting observations $y$ from $f$. The paper by \citet{jensen} provides an overview and comparison of these two methods; we review this below. For the subsequent discussion, we assume that we have a set of observations \REVISION{$y_i$} that satisfy the bound constraint: \REVISION{$a \le y_i \le b$}. 
\label{sec:transform_output}

\subsubsection{Warping Functions}
\label{sec:warping_fn}
Warping functions are used to transform bounded observations \REVISION{$y_i$} to unbounded
observations $u_i$. The field $u$ together with the observations $u_i$ are then treated with a traditional GP model using the steps outlined in Section \ref{gpr}.  The probit function, which is the inverse cumulative distribution function of a standard normal random variable: $\Phi^{-1}(\cdot)$, is commonly used as a warping function~\citep{jensen}. 
The probit function transforms bounded values \REVISION{$y\in[0,1]$} to unbounded values $u\in(-\infty,\infty)$ via
\begin{equation}\label{eq:warping_probit}
u = \Phi^{-1} \REVISION{\left(y\right)}
\end{equation}
The probit function is popular when \REVISION{$y_i$} is uniformly distributed in $[0,1]$ because the transformed values $u_i$ will be draws from a standard normal Gaussian with zero mean and unit variance. For a discussion of alternative warping functions we refer the reader to \citet{snelson2004warped}. 

\subsubsection{Likelihood formulations}
\label{sec:likelihood_formulations}
In addition to using warping functions, bound constraints can also be enforced using non-Gaussian likelihood functions $p(\bold{y}|X,\bold{f},\theta)$ that are constructed to produce GP observations which satisfy the constraints. Given a general non-Gaussian likelihood $p(\bold{y} | X,\bold{f},{\theta})$, the posterior distribution of GPR predictions is given by \eqref{eq:bayes}.
Unlike the posterior in \eqref{eq:gpreg}, the posterior in this case is no longer guaranteed to be Gaussian. There are a number of parametric distribution functions with finite support that can be used for the likelihood function to constrain the GP model.  \citet{jensen} suggest either a truncated Gaussian (see Section \ref{sec:mvn}) or the beta distribution scaled appropriately to the interval $[a,b]$.  Their results show that the beta distribution generally performs better.  

Unlike the warping method of Section \ref{sec:warping_fn}, with either a truncated Gaussian likelihood or a beta likelihood, the 
posterior \eqref{eq:bayes} is not analytically tractable. \citet{jensen} compare two schemes for 
approximate inference and prediction using bounded likelihood functions:  the Laplace approximation 
and expectation propagation.  These approaches both use a multivariate Gaussian approximation of the 
posterior, but solve for the governing posterior distribution in different ways. 

\subsection{Discrete Constraints using Truncated Gaussian Distributions}
\label{sec:daveiga}
By noting that a Gaussian process \eqref{eq:multivar_norm} is always trained and evaluated at a finite set of points $X$, global constraints over \REVISION{continuous domain $\Omega$ (such as an interval in one dimension)} can be approximated by constraints at a finite set of $N_c$ auxiliary or ``virtual'' points $\mathbf{x}_i, ..., \mathbf{x}_{N_c} \in \REVISION{\Omega}$. This approach, introduced by \citet{daveiga}, requires constructing an unconstrained GP and then, over the virtual points, transforming this GP to \REVISION{the} \emph{truncated} multivariate \REVISION{normal density $\mathcal{TN}(\textbf{z}; \bm{\mu},\Sigma,\textbf{a},\textbf{b})$ as a postprocessing step. The truncated multivariate normal is defined and discussed in detail in Section \ref{sec:mvn}.}
%\begin{equation}\label{eq:daveiga_tmn}
%\mathcal{TN}(\textbf{z}; \bm{\mu},\Sigma,\textbf{a},\textbf{b}) = 
%\begin{cases}
%\frac{\mathcal{N}(\textbf{z}; \bm{\mu},\Sigma)}{\mathbb{P}(\textbf{a}\leq\textbf{z}\leq\textbf{b})}, & \text{for  } \textbf{a}\leq\textbf{z}\leq\textbf{b} \\
%0, & \text{otherwise}
%\end{cases}
%\end{equation}
%as a postprocessing step. 

More specifically, \citet{daveiga} construct an approximation which is conditioned on a truncated multivariate Gaussian distribution at the auxiliary points.
We point out how this approach \REVISION{affects} the mean posterior predictions of the GP.
The unconstrained mean predictor is conditioned on the data $(X, \mathbf{y})$: 
\begin{equation}\label{eq:unconstrained_mean}
\mathbb{E}\left[ f(\bold{x}^*) \ \big| \ f(X) = \bold{y} \right].
\end{equation}
This setup is augmented by a fixed, finite set of discrete points $\{\mathbf{x}_i\}_{i=1}^{N_c}$, and the predictor \eqref{eq:unconstrained_mean} is replaced by the predictor 
\begin{equation}\label{eq:constrained_mean}
\mathbb{E}\left[ f(\bold{x}^*) \ \big| \ f(X) = \bold{y} \text{ and } a \le f(\mathbf{x}_i) \le b \text{ for all $i = 1, 2, ... N_c$}\right].
\end{equation}
As $[f(\bold{x}_1), ..., f(\bold{x}_{N_c})]^\top$ is normally distributed in the unconstrained case \eqref{eq:unconstrained_mean}, in the constrained case \eqref{eq:constrained_mean} it is distributed according to the truncated multivariate normal. 
 
In a few special cases, the mean and covariance of the truncated \REVISION{normal} can be derived analytically.  In one dimension, the mean at a 
single prediction point, $z_i$, is the unconstrained mean plus a factor which incorporates 
the change in the probability mass of the Gaussian distribution to reflect the truncation: 
\begin{equation}
\mathbb{E} \left({z_i} \mid  a\leq z_i\leq b \right) = \mu + \sigma\frac{\phi(\alpha)-\phi(\beta)}{\Phi(\beta)-\Phi(\alpha)}
\end{equation}
where $\alpha = \frac{a - \mu}{\sigma}$, $\beta = \frac{b-\mu}{\sigma}$, and $\phi$ and $\Phi$ are the probability density function and cumulative density function of a univariate standard normal distribution, respectively. In general, sampling and computing the moments of \REVISION{$\mathcal{TN}(\textbf{z}; \bm{\mu},\Sigma,\textbf{a},\textbf{b})$} is computationally demanding. \citet{daveiga} estimate moments empirically using an expensive rejection sampling procedure, based on a modified Gibbs sampler, to generate samples that honor the truncation bounds. We discuss the computational challenge of estimating the moments further in Section \ref{sec:mvn}. 

In contrast to the warping approach (Section~\ref{sec:transform_output}) or the spline approach (Section~\ref{sec:splines}) which maintain a global enforcement of the constraints, the bounds in \eqref{eq:constrained_mean} can depend on the location: $a_i \le f(\mathbf{x}_i) \le b_i$, representing different bounds in different regions of $I$ (see Section 4 of \cite{daveiga} for an example).
A downside of using the approach described here is that it is unclear how many virtual points $\bold{x}_i$ are needed to approximately constrain the GP globally with a prespecified level of confidence; some studies with increasing $N_c$ are presented by \citet{daveiga}. However, if the number of points can be chosen adequately, this approach can be used to enforce not only bound constraints but also monotonicity and convexity constraints~\citep{daveiga}; see Section \ref{sec:monotonicity_constraints} for more details. These types of constraints can also include linear transformations of a Gaussian process~\citep{agrell2019gaussian}.

\REVISION{
\subsection{Constrained maximum likelihood optimization to enforce nonnegativity constraints}
\label{sec:nonneg}
Another option for handling bound constraints is to constrain the optimization of the log-marginal-likelihood \eqref{eq:log_like}, so that hyperparameters are chosen to enforce bounds.  
\citet{pensoneault} introduced this approach to enforce nonnegativity constraints
up to a small probability $0 < \epsilon \ll 1$ of violation 
at a finite set of constraint points $\{\mathbf{x}_i\}_{i=1}^{N_c}$,
\begin{equation}
\label{eq:nonneg}
P\big( (f(\bold{x}^*_i) \ | \ \bold{y},X,\bold{x}^*_i,\bm{\theta}) < 0 \big) \leq \epsilon, \quad i = 1, 2, ..., N_c. 
\end{equation}
For a Gaussian likelihood, the unconstrained posterior $f^*$ follows a Gaussian distribution \eqref{eq:gpreg}, and the probabilistic constraint \eqref{eq:nonneg} can be written in terms of the posterior mean $\widehat{m}(\bold{x})$ and posterior standard deviation $s(\bold{x}) = \sqrt{\hat{v}(\bold{x})}$ given by \eqref{eq:post_mean_var}, and probit function $\Phi^{-1}$ (see Section \ref{sec:warping_fn}): 
\begin{equation}
\label{eq:nonneg2}
\widehat{m}(\bold{x}^*_i) + \Phi^{-1}(\epsilon)s(\bold{x}^*_i) \geq 0,  \quad i = 1, 2, ..., N_c. 
\end{equation}
\citet{pensoneault} chose $\epsilon = 2.3\%$ so that $\Phi^{-1}(\epsilon)=-2$, i.e., the mean minus two standard deviations 
is nonnegative. With the condition that $f(\bold{x}_j)$ be within $\nu > 0$ of the observations $y_j$, $j = 1, ..., N$, the maximization of the log-marginal-likelihood then becomes
\begin{align}
\begin{split}
\label{eq:nnlik}
\text{Seek }& \quad \bm{\theta}^* = \underset{\bm{\theta}}{\text{argmax}} \log(p(\bold{y}|X,\bm{\theta}))  \\
\textrm{subject to }& \quad  0  \leq \widehat{m}(\bold{x}_i)-2s(\bold{x}_i), \quad  i=1, ..., N_c \\  
\textrm{and }& \quad 0  \leq \nu - | y_j - f(\bold{x}_j)|,  \quad j=1, ..., N.  
\end{split}
\end{align}
\citet{pensoneault} solve the constrained optimization problem \eqref{eq:nnlik} with $\nu = 0.03$ using a nonlinear interior point solver, demonstrating that nonnegativity is enforced with high probability and also that posterior variance is significantly reduced.  
While this tends to be more expensive than a usual unconstrained optimization of the marginal log-marginal-likelihood, the effect on the posterior \eqref{eq:gpreg} is to change the hyperparameters, while preserving the Gaussian form, so that more expensive inference methods such as MCMC are not required. In principle, two-sided bounds and other types of constraints can be treated in this fashion, although \citet{pensoneault} consider nonnegative constraints in their numerical examples. 
}

\subsection{Splines}
\label{sec:splines}
\citet{maatouk} present a constrained Gaussian process formulation involving splines, where they place a multivariate Gaussian prior on a class of spline functions. The constraints are incorporated through constraints on the coefficients of the spline functions. 
To avoid the difficulty of enforcing a bound constraint $a \le f(\bold{x}) \le b$ globally on \REVISION{a continuous domain $\Omega$} for all predictions, the approach\REVISION{es} in Section\REVISION{s} \ref{sec:daveiga} \REVISION{and \ref{sec:nonneg}} enforced constraints only at a finite set \REVISION{of} points. 
In contrast, the approach taken by \citet{maatouk} is to instead consider a spline interpolant whose finite set of knot values are governed by a GP. This reduces the infinite-dimensional GP to a finite-dimensional one, for which the distributions of the knot values (i.e., the coefficients of the spline expansion) must be inferred. By using a set of piecewise linear splines that form a partition of unity, this approach guarantees that the set of all values between neighboring knots are bounded between the values of the knots. Thus if the knot values satisfy prescribed bound or monotonicity constraints, then so must all values in between them; that is, the global constraints are satisfied if the finite-dimensional constraints are. The problem then reduces to sampling the knot values from a truncated multivariate normal. 

\subsubsection{GPR for spline coefficients}

We first discuss the spline formulation in one input dimension, and without loss of generality assume that the process being modeled is restricted to the domain [0,1]. Let $h(x)$ be the standard tent function, i.e., the piecewise linear spline function defined by
\begin{equation}
h(x) = \text{max}(1-|x|,0)
\end{equation}
and define the locations of the knots to be $x_i = i/M$ for $i=0,1,...M$, with $M+1$ total spline functions. Then for any set of spline basis coefficients $\xi_i$, the function representation is given by
\begin{equation}\label{eq:spline_expansion}
f(x) = \sum_{i=0}^M \xi_i h(M(x-x_i)) = \sum_{i=0}^M \xi_i h_i(x).
\end{equation}
This function representation gives a $C^0$ piecewise linear interpolant of the point values $(x_i, \xi_i)$ for all $i=0,1,...,M$.

The crux of the spline approach to GPR lies in the following argument. Suppose we are given a set of $N$ data points at unique locations $(x_j,y_j)$. Define the matrix $A$ such that
\begin{equation}
A_{ij} = h_i(x_j).
\end{equation}
Then any set of spline coefficients $\bm{\xi}$ that satisfy the equation 
\begin{equation}\label{eq:spline_system}
A\bm{\xi} = \bold{y}
\end{equation}
will interpolate the data exactly. Clearly solutions to this system of equations will exist only if the rank of $A$ is greater than $N$, which requires that any given spline basis spans no more than two data points. Intuitively, this is because a linear function is only guaranteed to interpolate two points locally. Supposing that we make $M$ large enough to satisfy this condition, we can find multiple solutions to the system \eqref{eq:spline_system}.

\begin{figure}[htpb!]
\includegraphics[width=0.46\linewidth]{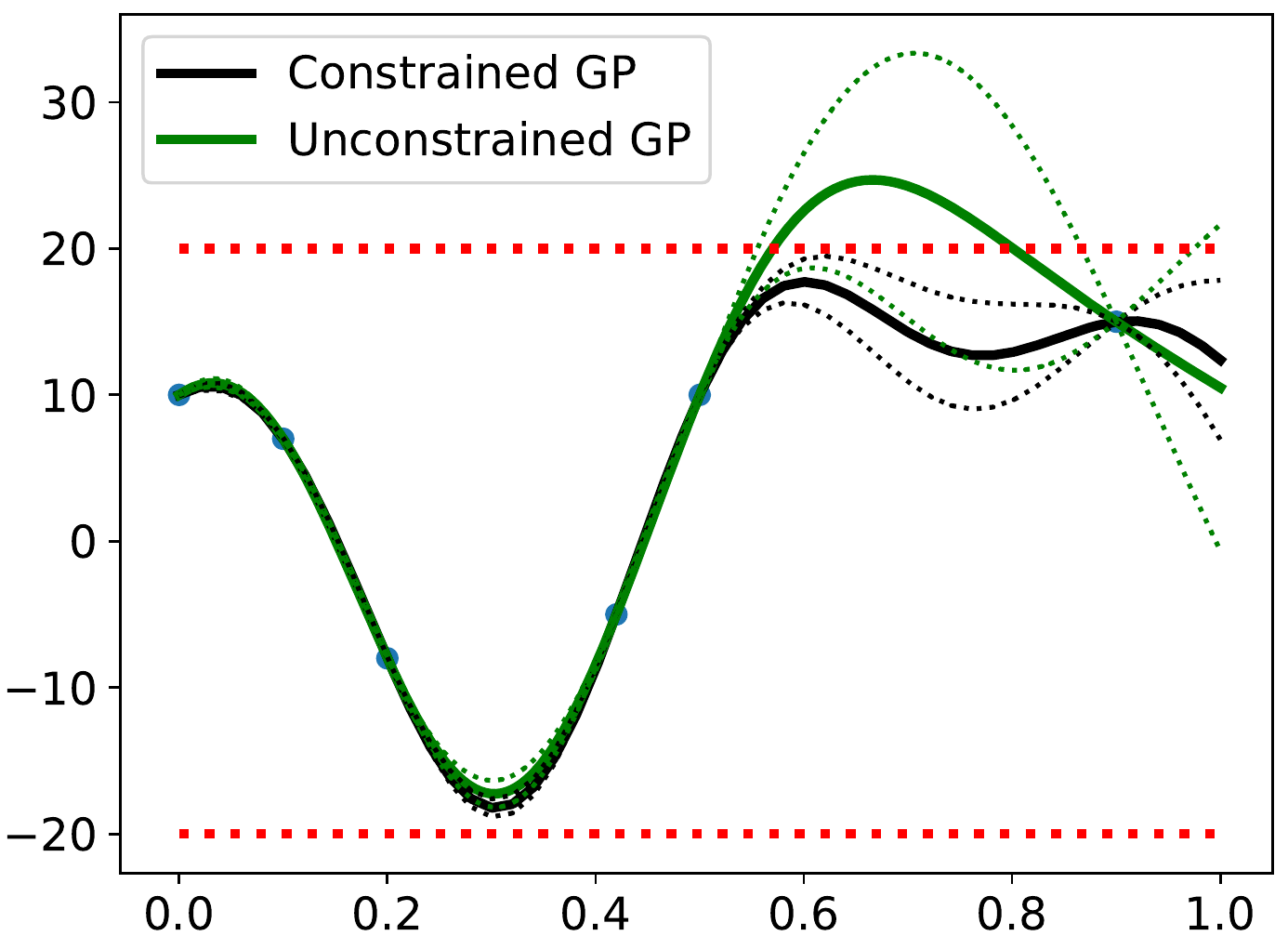}
\includegraphics[width=0.45\linewidth]{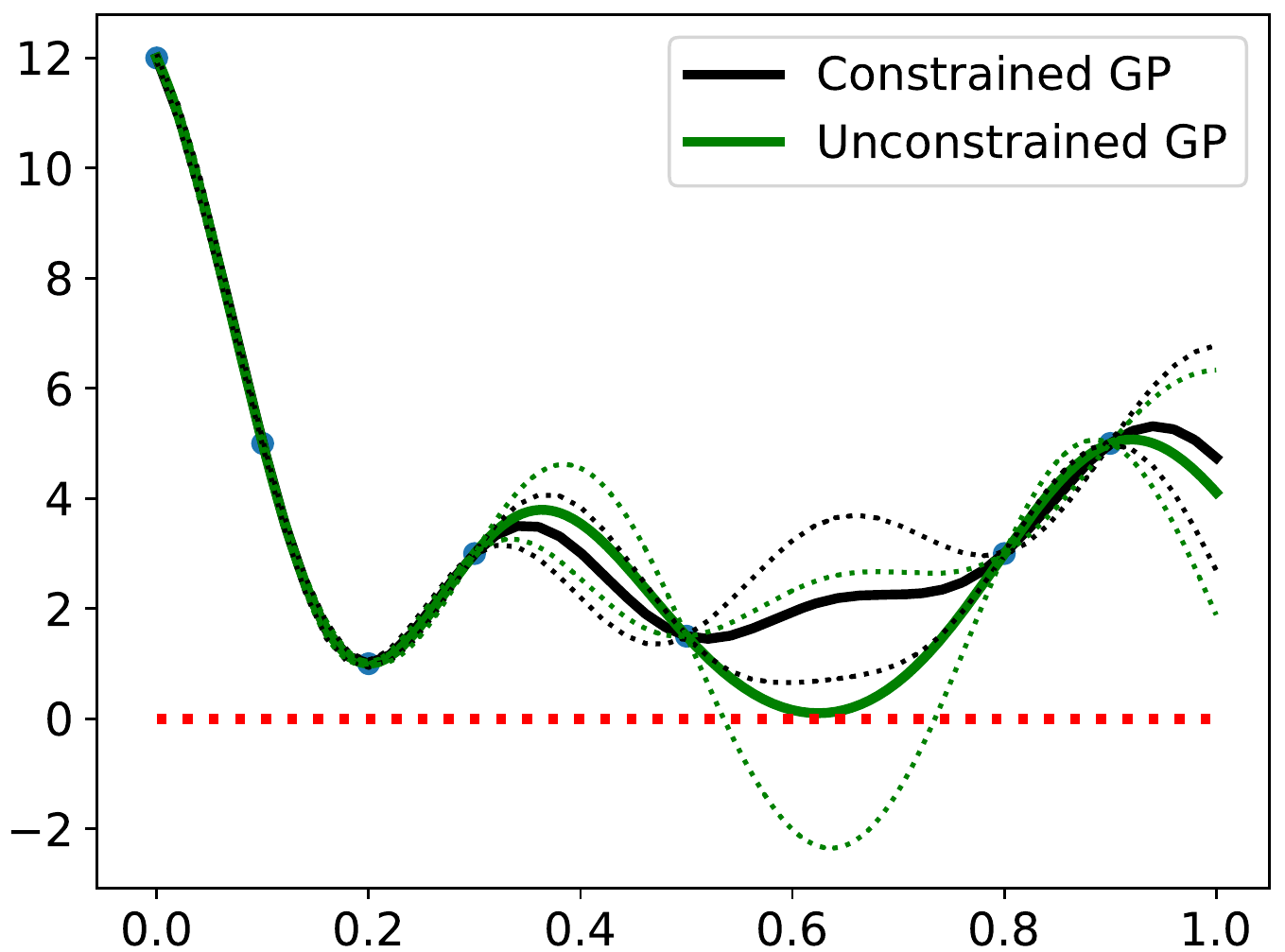}
\caption{
\emph{Left}: Comparison of a bound constrained GP with lower bound $a = -20$ and upper bound $b = 20$ versus an unconstrained GP. 
\emph{Right}: Comparison of a positivity constrained GP (lower bound $a = 0$) versus an unconstrained GP.
The data and hyperparameters are from \citet{maatouk}.
The dotted lines are $\mu \pm 1\sigma$ prediction intervals.
}\label{fig:fn12}
\end{figure}

We now assume the knot values $\bm{\xi}$ to be governed by a Gaussian process with covariance function $K$. Because a linear function of a GP is also a GP, the values of $\bm{\xi}$ and $\mathbf{y}$ are governed jointly \citep{maatouk,lopez2018} by a GP prior in the form 
\begin{equation}
\begin{bmatrix}
\mathbf{y} \\
\bm{\xi}
\end{bmatrix}
\sim \mathcal{N}\left(
\begin{bmatrix}
\mathbf{0}\\
\mathbf{0}
\end{bmatrix},
\begin{bmatrix}
AKA^\top & KA^\top\\
AK & K\\
\end{bmatrix}
\right)
\end{equation}
where each entry of the covariance matrix is understood to be a matrix. Upon observation of the data $\mathbf{y}$, the conditional distribution of the knot values subject to $\mathbf{y}=A\bm{\xi}$ is given by
\begin{equation}
p(\bm{\xi} \ \big| \ \mathbf{y}=A\bm{\xi}) = \mathcal{N}\Big(\bm{\xi}; KA^\top (AKA^\top)^{-1} \mathbf{y}, K - KA^\top(AKA^\top)^{-1}AK\Big)
\end{equation}
This formula is similar to that proposed by \citet{wilson}, in which a GP is interpolated to a regular grid design to take advantage of fast linear algebra. In this case, we are now interested in evaluating the distribution further conditioned on the inequality constraints given by
\REVISION{\begin{equation}\label{eq:spline_TN_C}
p(\bm{\xi} \ \big| \ \mathbf{y}=A\bm{\xi}, \mathbf{a} \leq \bm{\xi} \leq \mathbf{b}) = \mathcal{TN}\Big(\bm{\xi}; KA^\top (AKA^\top)^{-1} \mathbf{y}, K - KA^\top(AKA^\top)^{-1}AK ,\mathbf{a}, \mathbf{b} \Big)
\end{equation}
where %$\mathcal{C} = \{\bm{\xi} \ \big| \ \bm{a} \leq \bm{\xi} \leq \bm{b}\}$, and 
the truncated normal \REVISION{density} $\mathcal{TN}(\bm{\mu}, \Sigma, \mathbf{a}, \mathbf{b})$ is defined and discussed in Section \ref{sec:mvn}.} We illustrate bound constrained GPs using this approach in Figure \ref{fig:fn12}. We discuss monotonicity constraints using this approach in Section \ref{sec:splines_monotonic} and constrained MLE estimation of the hyperparameters in Section \ref{sec:mle}. Several constraint types can be combined in this approach, in which case $\mathcal{C}$ in \eqref{eq:spline_TN_C} is a convex set defined by a set of linear inequalities in $\bm{\xi}$ \citep{lopez2018}. 

\subsubsection{Sampling}
\label{sec:spline_sampling}
Just as for the discrete constraint method discussed in Section \ref{sec:daveiga}, sampling from the truncated normal distribution for the spline coefficients $\bm{\xi}$ introduces a new computational challenge into the GPR framework. While we discuss this in more detail and for several dimensions in Section \ref{sec:mvn}, we give a cursory discussion of this following \citet{maatouk}. We consider one dimension and the one-sided constraint $f(x) \ge b$ on $[0,1]$.

The original method of \citet{maatouk} was to use a rejection sampling approach by sampling from the untruncated 
distribution with a mean shifted to the mode (or maximum \emph{a posteriori} point) of the true posterior. That is, one first solves the problem
\begin{equation}\label{eq:posterior_mode}
\bm{\xi}^* = \underset{\bm{\xi}}{\text{argmin}}(\bm{\xi}-\bm{\mu})^\top\Sigma^{-1}(\bm{\xi}-\bm{\mu})
\end{equation}
subject to the bound constraints $\bm{\xi} \geq \mathbf{b}$,
where $\bm{\mu} = KA^\top (AKA^\top)^{-1} \mathbf{y}$ and $\Sigma = K-KA^\top(AKA^\top)^{-1}AK$. This is a convex quadratic program (assuming the covariance matrix is not too ill-conditioned) and may be solved efficiently.
One then draws samples from $\mathcal{N}(\bm{\xi}^*,\Sigma)$ and accepts or rejects the samples based on an inequality condition, described in more detail in \citet{maatouk_bay2016}.  
This is a simple approach, but it does not perform well at larger scale. The probability of rejecting any sample increases exponentially with the number of splines $M$. Furthermore, imprecision in the mode evaluation from the optimization process can lead to a deterioration of acceptance (for example, if the computed mode only satisfies monotonicity constraints to within some solver tolerance). 
Other approaches to sampling from the multivariate normal rely on Markov chain Monte Carlo methods, and are discussed in Section \ref{sec:mvn}.

\subsubsection{Multidimensional Setting}

The extension of the spline approach to higher dimensions is straightforward. The spline knots must be arranged in a regular grid with $M_j \REVISION{+ 1}$ points in each dimension \REVISION{$j$}, and the process is restricted to a hypercube domain of size $[0,1]^d$ for number of dimensions $d$. Under this restriction, the underlying function may be approximated with the \REVISION{tensor-product spline expansion}
\begin{equation}
\REVISION{
f(\mathbf{x}) = \sum_{i_1=0}^{M_{1}} \sum_{i_2=0}^{M_{2}} ... \sum_{i_d=0}^{M_{d}} \xi_{i_1, i_2, ..., i_d} h_{i_1, i_2, ..., i_d} 
(x_{1}, x_{2}, ..., x_{d}),
}
\end{equation}
\REVISION{where, with $h^j_{i} = h(M_j(x-x_{i_j}))$,
\begin{equation}
h_{i_1, i_2, ..., i_d}(\mathbf{x}) = h_{i_1}^1 \otimes h_{i_2}^2 \otimes ... \otimes h_{i_d}^d (\mathbf{x}) = h_{i_1}^1 (x_1) h_{i_2}^2(x_2) ... h_{i_d}^d(x_d),
\end{equation}}
for knot locations $(\REVISION{x_{i_1}, x_{i_2}, ..., x_{i_d}})$ and coefficients \REVISION{$\xi_{i_1, i_2, ..., i_d}$ for $0 \le i_j \le M_j$.} The inference process from this representation proceeds as before, as any set of observed data values can be expressed as $\bold{y}=A\bm{\xi}$ for the appropriately defined matrix $A$, and the coefficient values $\bm{\xi}$ may be inferred with a \REVISION{truncated multivariate normal distribution}. 

The primary issue with the multidimensional extension is the increase in cost. The spline approach suffers from the curse of dimensionality since the number of spline coefficients that must be inferred scales as $M^d$ with $M$ knots per dimension, leading to $O(M^{3d})$ scaling of the inference cost. This cost is further complicated by the fact that the spline formulation requires enough spline coefficients to guarantee interpolation through the data points in all dimensions, which means that $M\geq N$. Some potential methods for addressing computational complexity are discussed later in this work.
The need for efficient sampling schemes is also increased in the multidimensional setting as the acceptance ratio of a simple rejection sampler as discussed in Section \ref{sec:spline_sampling} decreases as the dimensionality (i.e. number of coefficients to infer) increases. This is partially addressed by the Gibbs sampling schemes referred to above, but those schemes also begin to lose efficiency as the size of the problem increases; for other approaches, see Section \ref{sec:mvn}.

\section{Monotonicity Constraints}\label{sec:monotonicity_constraints}
Monotonicity constraints are an important class of ``shape constraints'' which are frequently required in a variety of applications. For example, \citet{maatouk2017finite} applied monotonicity-constrained GPR for the output of the Los Alamos National Laboratory ``Lady Godiva'' nuclear reactor, which is known to be monotonic with respect to the density and radius of the spherical uranium core. \citet{kelly1990monotone} considered monotonic Bayesian modeling of medical dose-response curves, as did \citet{brezger2008monotonic} for predicting sales from various prices of consumer goods. 
 
Roughly speaking, given a method to enforce bound constraints, monotonicity constraints can be enforced by utilizing this method to enforce 
$\bold{f}' \ge \bold{0}$ on the derivative of the Gaussian process in a ``co-kriging'' setup for the joint GP 
$[\bold{f};\bold{f}']$. Indeed, many of the works reviewed in Section \ref{sec:bound_constraints} considered both bound, monotonicity, and convexity constraints under the general heading of ``linear inequality constraints'' \citep{maatouk,daveiga}. As a result, some of the methods below are based on techniques reviewed in Section \ref{sec:bound_constraints}, and we frequently refer to that section.

\subsection{Constrained Likelihood with Derivative Information}
\label{constrained_likelihood_with_derivative_information}

The work of \citet{riihimaki} enforces monotonicity of a Gaussian process using a probit model for the likelihood of the derivative observations.
Probit models are often used in classification problems or binary regression when one wants to predict a probability that a
particular sample belongs to a certain class (0 or 1)~\citep{rasmussen}.  
Here it is used generate a probability that the 
derivative is positive (1) or not (0). Monotonicity is obtained if the derivatives at all the selected points are 1 or 0. 

Using the probit model, the likelihood\footnote{This particular likelihood is the inverse of the probit function used for warping output in  \eqref{eq:warping_probit}:  
it maps a value from $(-\infty,\infty)$ to $[0,1]$, representing the probability that the value is in class 1 (which translates to monotonicity for this application).} for a particular derivative observation is given by 
\begin{equation}
\Phi(z) = \int_{-\infty}^{z} \REVISION{\mathcal{N}(t;0,1)} dt
\end{equation}
where \REVISION{$\mathcal{N}(t;0,1)$} is the probability density function of the standard one-dimensional normal distribution \eqref{standard_mvn}.  This likelihood is used within an expanded GPR framework that incorporates derivatives and constraints.
As part of this formulation, the original $n \times n$ GP covariance matrix, representing 
the covariance between $n$ data points, is extended to a
``four block'' covariance matrix. The full covariance 
matrix is composed of matrices involving the covariance between
function values, the covariance between derivative values, and the covariance
between function values and derivative values. 
 
Following \citet{riihimaki} our goal is to enforce the $d_i$-th partial derivative of $f$ at $\bold{x}_i$ to be nonnegative, i.e.
\begin{equation}\label{eq:discrete_monotonicity}
\frac{\partial f}{\partial x_{d_i}} (\bold{x}_i) \ge 0, 
\end{equation}
at a \REVISION{finite set of virtual points} $X_m=\{\bold{x}_i\}_{i=1}^m$. 
Using the shorthand notation
\begin{equation}\label{eq:notation_f}
f'_i = \frac{\partial f}{\partial x_{d_i}} (\bold{x}_i), \quad\text{and}\quad
\bold{f}' =
\begin{bmatrix}
\frac{\partial f}{\partial x_{d_1}}(\bold{x}_1) 
\hdots
\frac{\partial f}{\partial x_{d_m}}(\bold{x}_m) 	
\end{bmatrix}^\top
=
\begin{bmatrix}
f'_1
\hdots
f'_m	
\end{bmatrix}^\top
\end{equation}
and denoting\footnote{\citet{riihimaki} use the notation of $m_{d_i}^i$ rather than $y'_i$ for observations of ${\partial f}/{\partial x_{d_i}} (\bold{x}_i)$. \REVISION{They also use the term ``operating points'' for the virtual points.}}
an observation of $f'_i = {\partial f}/{\partial x_{d_i}} (\bold{x}_i)$ by $y'_i$,  we can write
\begin{equation}\label{eq:riihimaki_likelihood}
p\left(y'_i\big|f'_i \right) = \Phi\left(f'_i\frac{1}{\nu}\right).
\end{equation}
Here $\Phi(z)$ is the cumulative distribution function of the standard normal distribution and \eqref{eq:riihimaki_likelihood} approaches a step function as  $\nu \rightarrow 0$. 
Note that the likelihood function in ~\eqref{eq:riihimaki_likelihood} forces the likelihood to be zero (for non-monotonicity) or one (for monotonicity) in most cases. \REVISION{Therefore, by including observations of $y'_i = 1$ at the virtual points, the derivative is constrained to be positive}. \citet{riihimaki} point out that \eqref{eq:riihimaki_likelihood} is more tolerable of error than a step function, and use $\nu = 10^{-6}$ throughout their article. 

The joint prior is now given by: 
\begin{equation} 
p(\bold{f},\bold{f}'|X,X_m)=\mathcal{N}( \bold{f}_{\text{joint}} | \bm{0} , K_{\text{joint}} )
\end{equation} 
where
\begin{equation}\label{eq:likelihood_k_joint}%\label{eq:f_joint}
\bold{f}_{\text{joint}}=
\begin{bmatrix}
\bold{f} \\
\bold{f'} \\		
\end{bmatrix}
\quad \text{and} \quad
K_{\text{joint}}=
\begin{bmatrix}
K_{\bold{f,f}} &&  K_{\bold{f,f'}} \\
K_{\bold{f',f}} &&  K_{\bold{f',f'}} \\		
\end{bmatrix}.
\end{equation}
Here, the $n \times n$ matrix $K_{\bold{f,f}}$ denotes the standard covariance matrix for the GP $\bold{f}$ assembled over the data locations $X$: $K_{\bold{f,f}} = k(X,X)$, as in Section \ref{sec:gpr} where $k$ denotes the covariance function of $\bold{f}$. 
The $m \times m$ matrix $K_{\bold{f',f'}}$ in \eqref{eq:likelihood_k_joint} denotes the covariance matrix between the values of the specified partial derivatives of $\bold{f}$ at the operational points 
$X_m$:
\begin{equation}
\left[K_{\bold{f',f'}}\right]_{ij} = 
\left[
\text{cov}\left(
f'_i,
f'_j
\right)
\right]
=
\left[
\text{cov}\left(
\frac{\partial f}{\partial x_{d_i}}(\bold{x}_i),
\frac{\partial f}{\partial x_{d_j}}(\bold{x}_j)
\right)
\right], \quad
1 \le i,j \le m.
\end{equation}
\citet{riihimaki} show that $\frac{\partial f}{\partial x_{d_i}}$ is a GP with covariance matrix 
\begin{equation}\label{eq:derivative_covariance}
\frac{\partial}{\partial x_{d_i}} \frac{\partial}{\partial x'_{d_j}} k (\bold{x}, \bold{x}'),
\end{equation}
so that
\begin{equation}
\left[K_{\bold{f',f'}}\right]_{ij} = 
\frac{\partial^2  k }{\partial x_{d_i} \partial x'_{d_j}}(\bold{x}_i, \bold{x}'_j), \quad
1 \le i,j \le m.
\end{equation}
This result is a special case of a linear transformation of a GP; see Section \ref{pde_constraints} for more details. 
By the same general derivation in that section, 
the matrix $n \times m$ matrix $K_{\bold{f,f'}}$ represents the covariance between $\bold{f}$ and $\bold{f'}$, and is given by
\begin{equation}
\left[K_{\bold{f,f'}}\right]_{ij} = 
\frac{\partial  k }{\partial x'_{d_j}}(\bold{x}_i, \bold{x}'_j),
\quad
1 \le i \le n, 1 \le j \le m,
\end{equation}
and the $m \times n$ matrix $K_{\bold{f',f}} = K_{\bold{f,f'}}^\top$, representing the covariance between $\bold{f}'$ and $\bold{f}$.

\begin{figure}
\centering
\includegraphics[width=0.85\linewidth]{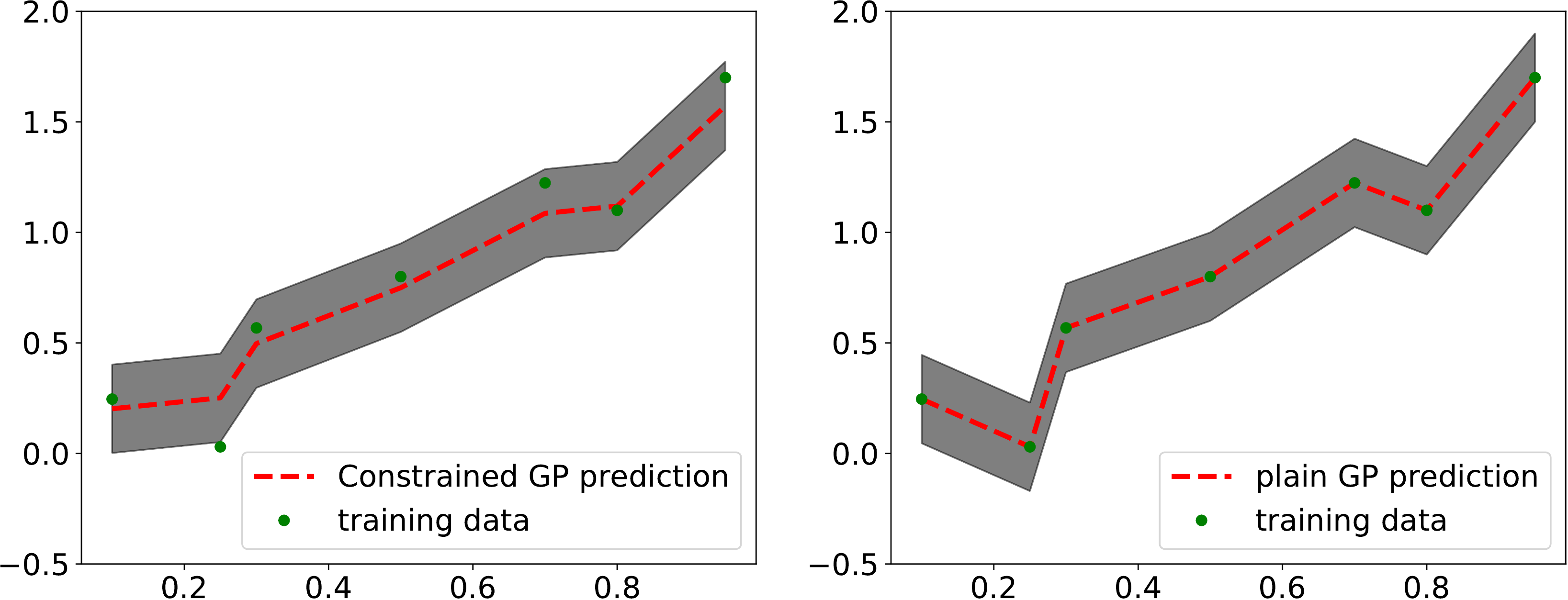}
\caption{Comparison of monotonic GP using the constrained likelihood formulation (\emph{left}) and the unconstrained GP (\emph{right}). \REVISION{Observations are given at $\{0.1, 0.25, 0.3, 0.5, 0.7, 0.8, 0.95\}$, with monotonicity constraints at the virtual points $\{0.15, 0.35, 0.55, 0.75\}$.}}
\label{fig:fn3}
\end{figure}

Putting this all together, we have the posterior probability \REVISION{density} of the joint distribution incorporating 
the derivative information\REVISION{:} 
\begin{equation}\label{riihimaki_posterior}
p(\bold{f},\bold{f}'|\bold{y},\bold{y}')=\frac{1}{Z}p(\bold{f},\bold{f}'|X,X_m)p(\bold{y}|\bold{f})p(\bold{y}'|\bold{f}')
\end{equation} 
where $1/Z$ is a normalizing constant.  This \REVISION{density} is analytically intractable
because of the non-Gaussian likelihood for the derivative components.  \citet{riihimaki} sample this \eqref{riihimaki_posterior} using expectation propagation. 
We used an MCMC approach to \REVISION{sample} the posterior distribution.  This approach is illustrated for an example in Figure \ref{fig:fn3}; this example is particularly challenging because the data is non-monotonic, but there is a requirement that the GP be monotonic. 

We describe the MCMC approach that we used for \eqref{riihimaki_posterior}. 
As before, $\bold{f}^*$ and $\bold{y}^*$ denote the estimates of these quantities at a new prediction point $\bold{x}^*$. 
\begin{equation}
p(\mathbf{y}^*|\mathbf{x}^*,\mathbf{x},\mathbf{y}) = \int p(\mathbf{y}^*|\mathbf{f}^*) p(\mathbf{f}^*|\mathbf{x}^*,\mathbf{x},\mathbf{y}) 
d\mathbf{f}^*,
\end{equation}
\begin{equation}
p(\mathbf{f}^*|\mathbf{x}^*,\mathbf{x},\mathbf{y}) = \int \int p(\mathbf{f}^*|\mathbf{x}^*,\mathbf{f},\mathbf{f}')p(\mathbf{f},\mathbf{f}'|\mathbf{x},\mathbf{y}) d\mathbf{f} d\mathbf{f}'.
\end{equation}
Since $p(\mathbf{f},\mathbf{f}'|\mathbf{x},\mathbf{y})$ was computed as samples from MCMC, we can approximate the posterior of $\mathbf{f}^*$ as
\begin{equation}
p(\mathbf{f}^*|\mathbf{x}^*,\mathbf{x},\mathbf{y}) = 
\REVISION{ \underset{\mathbf{f},\mathbf{f}' \sim p(\mathbf{f},\mathbf{f}'|\mathbf{x},\mathbf{y})} {\mathbb{E}}
\big[ p(\mathbf{f}^*|\mathbf{x}^*,\mathbf{f},\mathbf{f}') \big]} \approx \frac{1}{N}\sum_{i=1}^N p(\mathbf{f}^*|f_i,f'_i).
\label{eq:mcmc_monot}
\end{equation}
The MCMC samples outlined in \REVISION{\eqref{eq:mcmc_monot}} are generated over the 
vector $[\mathbf{f}; \mathbf{f}']$.   This could be a large vector, indicating a large latent function space, which may pose a challenge to MCMC.  Our experience indicates that one must start the MCMC sampling at a good initial point, one that is obtained either by finding the maximum \REVISION{\textit{a posteriori}} point (MAP) or by using a surrogate or interpolation to find a feasible initial point for $[\mathbf{f};\mathbf{f}']$. 

Note that this approach leads to the question of how to select the number and placement of the operating points $X_m$. \citet{riihimaki} point out that grid-based methods suffer from the curse of dimensionality, and that a more efficient strategy may be to successively add new operating points to $X_m$ by computing where derivatives of the GP are most likely to be negative for the current choice of $X_m$. 
We did not find much discussion of the placement of virtual points for this method or for the discrete constraint method in Section~\ref{sec:daveiga}.  The issue of optimal point placement for the virtual points could be addressed with 
some of the low-rank methods discussed in Section~\ref{sec:low_rank}.

\subsection{Monotonicity using Truncated Gaussian Distributions}
\label{daveiga_monotonic}
The approach, discussed in Section \ref{sec:daveiga}, for bounding Gaussian processes at a finite set of virtual points can naturally be extended to enforce monotonicity constraints. Specifically, by treating the partial derivatives ${\partial f}/{\partial x_{d_i}}$ as GPs with covariance kernel functions given by \eqref{eq:derivative_covariance}, monotonicity constraints of the same form as \eqref{eq:discrete_monotonicity} can be enforced at a discrete set of $N_c$ virtual points, i.e.
\begin{equation}
\frac{\partial f}{\partial x_{d_i}} (\bold{x}_i) \ge 0,\quad  i = 1, \ldots, N_c.
\end{equation}
This is done by treating the partial derivatives ${\partial f}/{\partial x_{d_i}}$ as GPs with covariance kernel functions given by \eqref{eq:derivative_covariance}, and using the joint Gaussian process $\mathbf{f}_{\text{joint}}$ with covariance matrix $\Sigma$ given by \eqref{eq:likelihood_k_joint}. 
Then, given data $(X,\bold{y})$ for $f$, 
\citet{daveiga} replace the unconstrained predictor \eqref{eq:unconstrained_mean} by the predictor
\begin{equation}\label{eq:derivative_daveiga_predictor}
\mathbb{E}\left[ f(\bold{x}^*) \ \Bigg| \ f(X) = \bold{y} \text{ and } 
0 \le \frac{\partial f}{\partial x_{d_i}}(\bm{x}_i) 
\text{ for all $i = 1, 2, ... N_c$}\right].
\end{equation}
This is analogous to the predictor \eqref{eq:constrained_mean} used for bound constraints. 
As a postprocessing step, per  \eqref{eq:derivative_daveiga_predictor} the \REVISION{density} $\mathcal{N}(\bm{\mu},\Sigma)$ for $\bold{f}'$ over the virtual points $\{\mathbf{x}_i\}_{i=1}^{N_c}$ is replaced by the \REVISION{density} $\mathcal{TN}(\bm{\mu},\Sigma,\bm{0},\bm{\infty})$; this \REVISION{density} is discussed more in Section \ref{sec:mvn}.

\subsection{Monotonic splines}
\label{sec:splines_monotonic}
The spline formulation, presented in Section \ref{sec:splines} to globally enforce a bound constraint of the form $f \ge a$  may be extended easily to enforce monotonicity constraints or other linear inequalities. For example, if $C$ is a first-order (backward or forward) finite difference matrix relating neighboring spline values, then monotonicity is enforced globally by sampling values of the knots $\bm{\xi}$ subject to the constraint
\begin{equation}
C\bm{\xi} \geq \mathbf{0};
\end{equation}
see \citet{lopez2018} or \citet{maatouk}.
This inequality is also used in the rejection sampler of Section \ref{sec:spline_sampling} as a constraint to identify the MAP estimate to increase the sampling efficiency.
Bound and monotonicity constraints can be enforced simultaneously by requiring both $\bm{\xi} \geq \mathbf{b}$ and $C\bm{\xi} \geq \mathbf{0}$ in the sampling, though the acceptance ratio drops substantially with combined constraints.

\subsection{Convexity}
\label{sec:convexity}
The sections above illustrated how a method for bound constraints can be used with first derivatives of a Gaussian process $f$ to enforce 
\REVISION{$\partial f/ \partial{x_{i}} \ge 0$} and thereby monotonicity for the GP $f$, either globally as in Section \ref{sec:splines_monotonic} or at a finite set of virtual points as in Sections \ref{constrained_likelihood_with_derivative_information} and \ref{daveiga_monotonic}. Similar nonnegativity constraints can be applied to higher derivatives of $f$ as well. In one dimension, this can be used to enforce convexity via the constraint
\begin{equation}\label{eq:one_dimension_convexity}
\frac{\partial^2 f}{\partial x^2} \ge 0,
\end{equation}
treating the left-hand side as a GP with covariance kernel 
\begin{equation}
\frac{\partial^4 k}{\partial x^2 \partial {x'}^2}(x,x').
\end{equation}

Although monotonicity can be enforced in arbitrary dimensions, convexity presents a challenge in dimensions greater than one, since it cannot be expressed as a simple linear inequality involving the derivatives of $f$ as in \eqref{eq:one_dimension_convexity}.
As \citet{daveiga} point out, enforcing convexity in higher dimensions requires that \eqref{eq:one_dimension_convexity} be replaced by the condition that the Hessian of $f$ be positive semidefinite. Sylvester's criterion yields the equivalent condition that each leading principal minor determinant of the Hessian be positive. Such inequality constraints involve \emph{polynomials} in partial derivatives of $f$. As polynomial functions of GPs are no longer GPs, the bound constraint methods in Section \ref{sec:bound_constraints} no longer apply. 

While higher dimensional convexity constraints are outside the scope of this survey, several references we have mentioned discuss the implementation of convexity constrained Gaussian processes in greater detail. 
\citet{daveiga} discuss how convexity in one dimension of the form \eqref{eq:one_dimension_convexity} can be enforced at virtual points
%$\bold{x}_i$, $i = 1, ..., N_c$
using the (partially) truncated multinormal, in a way analogous to Section \ref{daveiga_monotonic},
while convexity in two dimensions can be enforced using the elliptically truncated multinormal distribution.
%% This corresponds to the constrained predictor
%% \begin{equation}\label{eq:derivative_daveiga_2nd_predictor}
%% \mathbb{E}\left[ f(\bold{x}^*) \ \Bigg| \ f(X) = \bold{y} \text{ and } 
%% 0 
%% \le 
%% \frac{\partial^2 f}{\partial x_{d_i}^2}(\bm{x}_i) 
%% \text{ for all $i = 1, 2, ... N_c$}\right].
%% \end{equation}
\citet{maatouk} and \citet{lopez2018} point out that for the spline basis considered in Section \ref{sec:splines}, convexity in one dimension amounts to requiring that the successive differences of the values at the spline knots are increasing in magnitude, i.e. 
\begin{equation}
\xi_{k+1}-\xi_{k} \geq \xi_{k}-\xi_{k-1} \text{ for all } k.
\end{equation}
This is equivalent to requiring that the second-order finite differences be positive. This can also easily be applied in higher dimensions to guarantee that the second partial derivatives are positive globally, although this does not imply convexity.

\section{Differential Equation Constraints}\label{sec:pde_constraints}
Gaussian processes may be constrained to satisfy linear operator constraints of the form
\renewcommand\vec{\mathbf}
\begin{equation}
\label{eq:pde_constraint}
\mathcal{L} u = f
\end{equation}
given data on $f$ and $u$. 
When $\mathcal{L}$ is a linear partial differential operator of the form
\begin{equation}
\label{eq:lin_diff_op}
\mathcal{L} = \sum_{\bm{\alpha}} C_{\bm{\alpha}}(\mathbf{x}) \frac{\partial^{\bm{\alpha}}}{\partial \mathbf{x}^{\bm{\alpha}}}, \quad
\bm{\alpha} = (\alpha_1, ..., \alpha_d), \quad
\frac{\partial^{\bm{\alpha}}}{\partial \mathbf{x}^{\bm{\alpha}}} = 
\frac{\partial^{{\alpha}_1}}{\partial {x}_1^{{\alpha}_1}}
\frac{\partial^{{\alpha}_2}}{\partial {x}_2^{{\alpha}_2}}
...
\frac{\partial^{{\alpha}_d}}{\partial {x}_{\REVISION{d}}^{{\alpha}_d}},
\end{equation}
the equation \eqref{eq:pde_constraint} can be used to 
constrain GP predictions to satisfy known physical laws expressed as linear partial differential equations.
In this section we survey methods to constraint GPs with PDE constraints of the form \eqref{eq:pde_constraint}.

\subsection{Block Covariance Kernel} 
\label{pde_constraints}
\REVISION{The work of \citet{raissi2017} introduced a joint GPR approach using a four-block covariance kernel allowing observations of both the solution $u$ and the forcing $f$ to be utilized. The principle behind this approach} is that if $u(\mathbf{x})$ is a GP with mean function $m(\mathbf{x})$ and covariance kernel
$k(\mathbf{x},\mathbf{x'})$, 
\begin{equation}
\label{eq:gp_for_u}
u \sim \mathcal{GP}(m(\mathbf{x}),k(\mathbf{x},\mathbf{x'}))
\end{equation}
and if $m(\cdot)$ and $k(\cdot, \mathbf{x'})$ belong to the domain of $\mathcal{L}$, then
$\mathcal{L}_{\mathbf{x}} \mathcal{L}_{\mathbf{x'}} k(\mathbf{x},\mathbf{x'})$ defines a valid covariance kernel for a 
GP with mean function $\mathcal{L}_{\mathbf{x}} m(\mathbf{x})$. 
This Gaussian process is denoted $\mathcal{L}u$:
\begin{equation}
\label{eq:gp_for_Lu}
\mathcal{L}u \sim \mathcal{GP}(\mathcal{L}_{\mathbf{x}} m(\mathbf{x}), \mathcal{L}_{\mathbf{x}} \mathcal{L}_{\mathbf{x'}} k(\mathbf{x},\mathbf{x'})).
\end{equation}
Note from \eqref{eq:lin_diff_op} that the operator $\mathcal{L}$ takes as input a function of a single variable $\mathbf{x}$. When applying 
$\mathcal{L}$ to a function of two variables such as $k(\mathbf{x}, \mathbf{x'})$, we use a subscript as in \eqref{eq:gp_for_Lu} to denote the application of $\mathcal{L}$ in the indicated variable, i.e., considering the input to $\mathcal{L}$ as a function of the indicated variable only. Note that a special case of this, for $\mathcal{L} = \partial/\partial x_{d_i}$, appeared in Section \ref{constrained_likelihood_with_derivative_information}. The same formula \eqref{eq:gp_for_Lu} was utilized in earlier works on GPR with differential equation constraints by \citet{graepel2003solving} and \citet{sarkka2011linear}. \citet{owhadi2015bayesian} showed a similar result in addressing the problem of identification of basis elements in the finite-dimensional approximation of PDE solutions.  The latter work presents a Bayesian formulation of a numerical homogenization approach in which the optimal bases are shown to be polyharmonic splines and the optimal solution in the case of a white noise source term is a Gaussian field.

The notation ``$\mathcal{L}u$'' for the GP \eqref{eq:gp_for_Lu} is suggested by noting that if one could apply $\mathcal{L}$ to the samples of the GP $u$, then the mean
of the resulting stochastic process $\mathcal{L}[u]$ would indeed be given by 
\begin{align}
\text{mean}\left(\mathcal{L}[u](\mathbf{x})\right) &=
\mathbb{E} \left[ \mathcal{L} [u](\mathbf{x}) \right] = \mathcal{L} \mathbb{E}\left[ u(\mathbf{x}) \right] = \mathcal{L} m(\mathbf{x})
\end{align}
and the covariance by
\begin{align}
\begin{split}
\label{eq:cov_f_f}
\text{cov}\left( \mathcal{L}[u](\mathbf{x}) , \mathcal{L}[u](\mathbf{x'}) \right) &= 
\mathbb{E} \left[ \mathcal{L}_{\mathbf{x}} [u(\mathbf{x})] \mathcal{L}_{\mathbf{x'}} [u (\mathbf{x'})] \right] =
\mathbb{E} \left[ \mathcal{L}_{\mathbf{x}} \mathcal{L}_{\mathbf{x'}} \left[ u(\mathbf{x}) u(\mathbf{x'}) \right] \right] \\
\qquad &=  \mathcal{L}_{\mathbf{x}}  \mathbb{E} \left[ \mathcal{L}_{\mathbf{x'}} \left[ u(\mathbf{x}) u(\mathbf{x'}) \right] \right] 
= \mathcal{L}_{\mathbf{x}} \mathcal{L}_{\mathbf{x'}} \mathbb{E} \left[ u(\mathbf{x}) u(\mathbf{x'}) \right] \\
&= \mathcal{L}_{\mathbf{x}} \mathcal{L}_{\mathbf{x'}}  \left[ \text{cov}\left( u(\mathbf{x}), u(\mathbf{x'}) \right) \right] 
= \mathcal{L}_{\mathbf{x}} \mathcal{L}_{\mathbf{x'}}  k(\mathbf{x}, \mathbf{x'}).
\end{split}
\end{align}
This justification is formal, as in general the samples of the process $\mathcal{L}u$ defined by \eqref{eq:gp_for_Lu} cannot be identified as $\mathcal{L}$ applied to the samples of $u$ \citep{kanagawa2018gaussian,driscoll1973reproducing}; a rigorous interpretation involves the posterior predictions and reproducing kernel Hilbert spaces of the processes $u$ and $\mathcal{L}u$ \citep{seeger2004gaussian,berlinet2011reproducing}.

If scattered measurements
$\mathbf{y}_f$ %$Y_f = \{\mathbf{y}_f\}$
on the source term $f$ in \eqref{eq:pde_constraint} are available at domain points $X_f$,
% = \{\mathbf{x}_f\}$
then this can be used to train and obtain predictions for $\mathcal{L}u$ from the GP \eqref{eq:gp_for_Lu}  in the standard way. If, in addition, measurements $\mathbf{y}_u$
%$Y_u = \{\mathbf{y}_u\}$
of $u$ are available at domain points $X_u$
% = \{\mathbf{x}_u\}$, 
a GP co-kriging procedure can be used. In this setting physics knowledge of the form \eqref{eq:pde_constraint} enters via the data $(X_f, \mathbf{y}_f)$ and can be used to improve prediction accuracy and reduce variance of the GPR of $u$. 
The co-kriging procedure requires forming the joint Gaussian process $[u;f]$.
%$
%\begin{bmatrix}
%u \\ f
%\end{bmatrix}
%$.
Similarly to the derivative case considered in Section \ref{constrained_likelihood_with_derivative_information}, the covariance matrix of the resulting GP is a four block matrix assembled from
the covariance matrix of the GP \eqref{eq:gp_for_u} for the solution $u$, the covariance of the GP \eqref{eq:gp_for_Lu} for the forcing function, and the cross terms.
Given the covariance kernel $k(\mathbf{x}, \mathbf{x}')$ for $u$, the covariance kernel of this joint GP 
is
\begin{equation}\label{eq:joint_covariance}
k
\left(
\begin{bmatrix} 
\mathbf{x}_1 \\ 
\mathbf{x}_2 
\end{bmatrix}
,
\begin{bmatrix} 
\mathbf{x}'_1 \\ 
\mathbf{x}'_2 
\end{bmatrix}
\right)
=
\begin{bmatrix}
\phantom{\mathcal{L}_{\mathbf{x}}} k(\mathbf{x}_1,\mathbf{x}'_1) & 
\phantom{\mathcal{L}_{\mathbf{x}}}\mathcal{L}_{\mathbf{x}'}  k(\mathbf{x}_1,\mathbf{x}'_2)\\ 
\mathcal{L}_{\mathbf{x}}   k (\mathbf{x}_2,\mathbf{x}'_1) 
& \mathcal{L}_{\mathbf{x}} \mathcal{L}_{\mathbf{x'}} k(\mathbf{x}_2,\mathbf{x}'_2)
\end{bmatrix}
=
\begin{bmatrix}
K_{11} & 
K_{12} \\ 
K_{21} &
K_{22} 
\end{bmatrix}.
\end{equation}
The covariance between $u(\mathbf{x})$ and $f(\mathbf{x'})$ is given by $\mathcal{L}_{\mathbf{x}'}  k(\mathbf{x}_1,\mathbf{x}'_2)$ in the upper right block of the kernel and can be justified by a calculation similar to \eqref{eq:cov_f_f}; see \citet{raissi2017}. Similarly the covariance between $u(\mathbf{x}')$ and $f(\mathbf{x})$ is represented by the bottom left block $\mathcal{L}_{\mathbf{x}} k(\mathbf{x}_2,\mathbf{x}'_1)$ of the kernel. 
In this notation, the joint Gaussian process for $[u; f]$
%$
%\begin{bmatrix}
%u \\ f
%\end{bmatrix}
%$
is then
\begin{equation}\label{eq:joint_GP_uf}
\begin{bmatrix}
u({X}_1) \\ 
f({X}_2)
\end{bmatrix}
\sim \mathcal{GP}\left( 
\begin{bmatrix}
\phantom{\mathcal{L}}
m({X}_1) \\
\mathcal{L}
m({X}_2)
\end{bmatrix},
\begin{bmatrix}
K_{11}(X_1,X_1) & 
K_{12}(X_1,X_2) \\ 
K_{21}(X_2,X_1) &
K_{22}(X_2,X_2)
\end{bmatrix}
\right),
\end{equation}
where $K_{12}(X_1,X_2) = \left[K_{21}(X_2,X_1)\right]^\top$.

Given data $({X}_u,\mathbf{y}_u)$ and $({X}_f,\mathbf{y}_f)$, the GP kernel hyperparameters 
may be trained by assembling the four-block covariance matrix in \eqref{eq:joint_GP_uf} with 
${X}_1 = {X}_u$, ${X}_2 = {X}_f$,
\begin{equation}\label{eq:covariance_over_data}
K_{\text{data}}=
\begin{bmatrix}
K_{11}({X}_u,{X}_u) & 
K_{12}({X}_u,{X}_f) \\ 
K_{21}({X}_f,{X}_u) &
K_{22}({X}_f,{X}_f)
\end{bmatrix}
\end{equation}
and minimizing the negative log-marginal-likelihood
\begin{multline}
\label{NLML}
-\log{p(\bold{y}_u,\bold{y}_f|X_u,X_f,\bm{\theta)}} = 
\frac{1}{2} \left(
\mathbf{y}-\mathbf{m}
\right)^{\top}
K_{\text{data}}^{-1}
\left(
\mathbf{y}-\mathbf{m}
\right)
 + \frac{1}{2}\log |{K}_{\text{data}}| + \frac{N}{2} \log (2\pi), \\
\text{with } 
\mathbf{y} 
= 
\left[
\begin{array}{c}
\mathbf{y}_u \\ 
\mathbf{y}_f
\end{array} 
\right]
\text{ and }
\mathbf{m} 
=
\left[
\begin{array}{c}
\phantom{\mathcal{L}} m({X}_u) \\ 
\mathcal{L} m({X}_f)
\end{array} 
\right].
\end{multline} 
In the presence of noise on measurements of $u$ and $f$, a standard approach analogous to the Gaussian likelihood \eqref{eq:gauss_like} is to introduce two noise hyperparameters $\sigma_{u}$ and $\sigma_{f}$ and replace the four-block covariance
matrix \eqref{eq:covariance_over_data} by 
\begin{equation}
\begin{bmatrix*}[l]
K_{11}({X}_u,{X}_u)          + \sigma_{u}^2 {I}_{N_u}& 
K_{12}({X}_u,{X}_f)\\ 
K_{21}({X}_f,{X}_u)&
K_{22}({X}_f,{X}_f)          + \sigma_{f}^2 {I}_{N_f}
\end{bmatrix*}
\end{equation}
The inclusion of the additional terms depending on $\sigma_{u}^2$ and $\sigma_{f}^2$ correspond to an assumption of uncorrelated 
white noise on the measurements ${Y}_u$ and ${Y}_f$, i.e., 
\begin{equation}
{Y}_u = u({X}_u) + \bm{\epsilon}_u, \quad
{Y}_f = f({X}_f) + \bm{\epsilon}_f,
\end{equation}
with $\bm{\epsilon}_u \sim \mathcal{N}(\mathbf{0},\sigma_{u}^2 {I}_{N_u})$ and independently $\bm{\epsilon}_f \sim \mathcal{N}(\mathbf{0},\sigma_{f}^2 {I}_{N_u})$, given $N_u$ data points for $u$ and $N_f$ data points for $f$. 

The implementation of the constrained Gaussian process kernel \eqref{eq:joint_covariance} for constraints of the form
\eqref{eq:pde_constraint} raises several computational problems.
The first is the computation of $\mathcal{L}_{\mathbf{x}}k$ and $\mathcal{L}_{\mathbf{x}} \mathcal{L}_{\mathbf{x'}}k$. The most ideal scenario is that in which $k$ has an analytical formula and $\mathcal{L}$ is a linear differential operator so that these expressions be computed in closed form by hand or with a symbolic computational software such as Mathematica. This was the approach used for the examples in \citet{raissi2017}, \citet{raissi2018numerical} and \citet{raissi2018}, including for the heat equation, Burgers' equation, Korteweg-de Vries Equation, and Navier-Stokes equations. The nonlinear PDEs listed here were treated using an appropriate linearization. An example of $k$ being parametrized by a neural network (which allows derivatives to be computed using backpropagation) was also considered in \citet{raissi2018numerical} for the Burgers' equation. 

Closed form expressions for the covariance kernel \eqref{eq:joint_covariance} greatly simplify the implementation compared to numerical approximation of $\mathcal{L}_{\mathbf{x}}k$ and $\mathcal{L}_{\mathbf{x}} \mathcal{L}_{\mathbf{x'}}k$ using finite-differences or series expansions. As the size of the dataset and therefore size of the covariance matrix \eqref{eq:joint_covariance} increases, our numerical experiments suggest that any numerical errors in the approximation of the action of $\mathcal{L}$ rapidly lead to ill-conditioning of the covariance matrix. This in turn can lead to artifacts in the predictions or failure of maximum likelihood estimation with the constrained GP. Ill-conditioning can be reduced by adding an ad-hoc regularization on the diagonal of \eqref{eq:joint_covariance} at the cost of reducing the accuracy of the regression, potentially negating the benefit of the added constraint.
For more general constraints of the form \eqref{eq:pde_constraint}, depending on the form of $k$ or $\mathcal{L}$, numerical methods may be unavoidable. For example, in \citet{raissi2017} and \citet{gulian2019machine}, fractional-order PDE constraints (amounting to $\mathcal{L}$ being a nonlocal integral operator with singular kernel) were considered. For these constraints, the kernel blocks $\mathcal{L}_{\mathbf{x}}k$ and $\mathcal{L}_{\mathbf{x}} \mathcal{L}_{\mathbf{x'}}k$ had no closed formula. To approximate these terms, a series expansion was used in \citet{raissi2017}, and in \citet{gulian2019machine} a numerical method was developed involving Fourier space representations of $\mathcal{L}_{\mathbf{x}}k$ and $\mathcal{L}_{\mathbf{x}} \mathcal{L}_{\mathbf{x'}}k$ with Gaussian quadrature for Fourier transform inversion. 

A second problem is that the formulation \eqref{eq:joint_GP_uf} requires enforcing the constraint \eqref{eq:pde_constraint} at discrete points of ${X}_f$. Therefore, even if we have complete knowledge of the constraining equation \eqref{eq:pde_constraint} and the forcing term $f$, enhancing the GPR for $u$ by including a high number of virtual data points makes inference as well as maximum likelihood estimation computationally expensive and prone to ill-conditioning. In this regard, the computational approaches discussed in Section \ref{sec:low_rank}, particularly the subset of data approaches in Section \ref{sec:subset_of_data}, may be helpful.

\begin{figure}
  \centering
  \includegraphics[width=\textwidth]{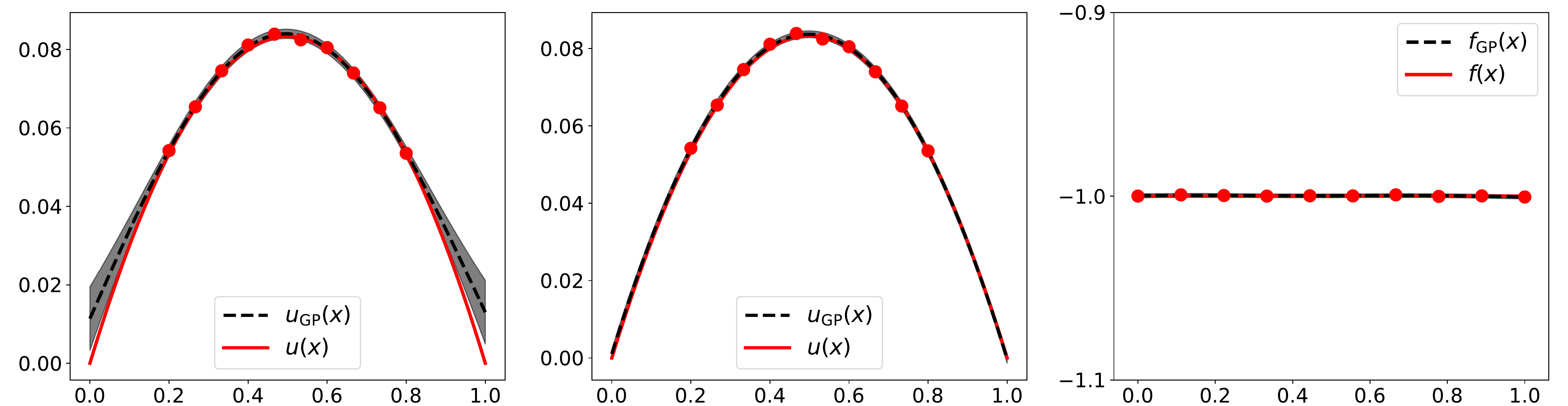}
  \caption{Comparison of unconstrained and PDE constrained GP. \emph{Left:} Reconstruction of $u$ (red line) with an unconstrained GP (black line) using 10 data points (red dots) in $[0.2, 0.8]$. \emph{Center:} Reconstruction of $u$ (red line) with a PDE constrained GP (black line) using the same 10 data points (red dots) in $[0.2, 0.8]$. \emph{Right:} Right-hand side $f$ of the PDE, with 10 additional data points in $[0,1]$ used for the PDE constraint. Note the improved accuracy of the constrained GP outside $[0.2, 0.8]$ due to this constraint data.}
  \label{fig:pde_example_1}
\end{figure}

%\subsubsection{Example} 
%\label{pde_examples}
Figure \ref{fig:pde_example_1} shows an example of a one-dimensional GP with squared-exponential kernel constrained to satisfy the differential equation $1 = d^2 u / dx^2$ on the interval $[0,1]$. Data is generated from sampling the solution $u = \frac{1}{8} [(2x-1)^2-1]$ at 10 points between 0.2 and 0.8. Both the constrained and unconstrained GPs give a reasonably accurate reconstruction on $[0.2, 0.8]$, but the unconstrained GP has poor accuracy outside this subinterval. On the other hand, the constrained GP is augmented by data $f = d^2 u / dx^2 = 1$ at 10 additional points between $0$ and $1$, leading to an improved reconstruction of $u$ outside $[0.2,0.8]$. 

\subsection{Transformed Covariance Kernel}
\label{transformed_covariance}
A different approach to constrain Gaussian processes by differential equations is to design a specialized covariance kernel such that the GP satisfies the constraint globally, rather than at a discrete set of auxiliary data points as in Section \ref{pde_constraints}. This method dates back to the divergence-free kernel of \citet{narcowich1994generalized} for vector-valued GPs. In addition to being a stronger enforcement of the constraint, this method also avoids the computational burden induced by the four-block covariance matrix. On the other hand, it has more limited applicability and specialized implementation, as it requires analytically solving for a kernel with the desired constraining property. The authors of \citet{jidling2017linearly} propose finding a linear operator $\mathcal{G}$ which maps a certain class of functions (modeled by a GP) to the null space of the linear differential operator defining the constraint. The operator $\mathcal{G}$ can then used to compute the constrained kernel as a transformation of a starting kernel. We now summarize this approach, provide some examples, and compare it in greater detail to other approaches. 

Given a linear operator $\mathcal{L}_{\mathbf{x}}$ and a vector-valued GP $\mathbf{f}$ described using a matrix-valued covariance kernel function that encodes the covariance between the entries of the vector $\mathbf{f}$, the constraint
\begin{equation}\label{transformed_constraint}
\mathcal{L}_{\mathbf{x}}\mathbf{f} = 0
\end{equation}
is satisfied if $\mathbf{f}$ can be represented as
\begin{equation}\label{f_from_g}
\mathbf{f} = \mathcal{G}_{\mathbf{x}}\mathbf{g},
\end{equation}
for a transformation $\mathcal{G}_{\mathbf{x}}$ such that
\begin{equation}\label{operator_equation}
\mathcal{L}_{\mathbf{x}} \mathcal{G}_{\mathbf{x}} = 0.
\end{equation}
In other words, the range of the operator $\mathcal{G}_{\mathbf{x}}$ lies in the nullspace of the operator $\mathcal{L}_{\mathbf{x}}$.
Further, provided that
$\mathcal{G}_{\mathbf{x}}$ is also a linear operator, if $\mathbf{g}$ is a GP with covariance kernel $k_{\mathbf{g}}$, then by  \eqref{f_from_g} $\mathbf{f}$ is also a GP with covariance kernel
\begin{equation}\label{transformed_kernel}
k_{\mathbf{f}} = \mathcal{G}_{\mathbf{x}} k_{\mathbf{g}} \mathcal{G}_{\mathbf{x'}}^\top.   
\end{equation}
Above and throughout this section, we follow \citet{jidling2017linearly} in using the notation $\mathcal{G}_{\mathbf{x}} k_{\mathbf{g}} \mathcal{G}_{\mathbf{x'}}^\top$ for the matrix-valued function with $(i,j)$-entry $\REVISION{\sum_{k} \sum_{l}}\left[ \mathcal{G}_{\mathbf{x}} \right]_{ik}
\left[ \mathcal{G}_{\mathbf{x'}} \right]_{jl} \left[ k_{\mathbf{g}}(\mathbf{x}, \mathbf{x}') \right]_{kl}$; 
note that if $\mathbf{g}$ and therefore $k_{\mathbf{g}}$ are scalar valued, this reduces to \eqref{eq:cov_f_f}.   
If the operator equation \eqref{operator_equation} can be solved, one can choose a kernel $k_{\mathbf{g}}$ and define a GP $\mathbf{f}$ using \eqref{transformed_kernel} which satisfies the constraint \eqref{transformed_constraint}. The constraint is satisfied globally by the structure of the covariance kernel; no data is required to enforce it. We refer to this as the \emph{transformed covariance kernel} approach. 
Prototypical examples applying the constraint \eqref{transformed_constraint} are divergence-free and curl-free constraints on the vector field $\mathbf{f}$; an excellent illustration of such GP vector fields is given by \citet{macedo2010learning}.
%\REVISION{\citet{Albert} also develop specialized kernels and provide a valuable set of examples for Laplace's equation, the heat equation, and the Helmholtz equation. 
%Their approach involves a Gaussian process constructed with specialized kernels which generate an exact solution of the homogeneous equation.  The inhomogeneous contributions 
%involving linear superposition of point sources are handled with a linear model.}

As an example of how to apply \eqref{transformed_constraint}, consider the enforcement of the curl-free constraint 
\begin{equation}
\mathcal{L}_{\mathbf{x}} f = 
\nabla \times \mathbf{f} = 0
\end{equation}
for a vector field $\mathbf{f}: \mathbb{R}^3 \rightarrow \mathbb{R}^3$. A curl-free vector field can be written
$\mathbf{f} = \nabla g$,
for a scalar function $g : \mathbb{R}^3 \rightarrow \mathbb{R}$. So, for $\mathcal{L}_{\mathbf{x}} = \nabla \times$ the choice 
\begin{equation}
\mathcal{G}_{\mathbf{x}}  = \nabla, \quad \text{i.e.} \quad
\mathcal{G}_{\mathbf{x}} g = 
\begin{bmatrix}
\frac{\partial g}{\partial x_1} \\[6pt]
\frac{\partial g}{\partial x_2} \\[6pt]
\frac{\partial g}{\partial x_3}
\end{bmatrix}
=
\begin{bmatrix}
\frac{\partial}{\partial x_1} \\[6pt]
\frac{\partial}{\partial x_2} \\[6pt]
\frac{\partial}{\partial x_3}
\end{bmatrix}
g
\end{equation}
satisfies \eqref{operator_equation}. 
Thus, placing a GP with scalar-valued covariance kernel $k_g(\mathbf{x},\mathbf{x}')$ on $g$ leads via \eqref{transformed_kernel} to a $3 \times 3$ matrix-valued covariance kernel 
\begin{equation}
k_{\text{curl-free}}(\mathbf{x},\mathbf{x}') = 
\mathcal{G}_{\mathbf{x}}  k_g \mathcal{G}_{\mathbf{x}}^\top 
= 
\begin{bmatrix*}
\frac{\partial^2 }{\partial x_1 \partial x_1'}
&
\frac{\partial^2 }{\partial x_1 \partial x_2'}
&
\frac{\partial^2 }{\partial x_1 \partial x_3'}
\\[6pt]
\frac{\partial^2 }{\partial x_2 \partial x_1'}
&
\frac{\partial^2 }{\partial x_2 \partial x_2'}
&
\frac{\partial^2 }{\partial x_2 \partial x_3'}
\\[6pt]
\frac{\partial^2 }{\partial x_3 \partial x_1'}
&
\frac{\partial^2 }{\partial x_3 \partial x_2'}
&
\frac{\partial^2 }{\partial x_3 \partial x_3'}
\end{bmatrix*}
k_g (\mathbf{x}, \mathbf{x}').
\end{equation}
Here, $k_g$ is a scalar-valued kernel. 
If the squared-exponential covariance kernel 
$k_g(\mathbf{x},\mathbf{x}') = \gamma e^{-\frac{|\mathbf{x} - \mathbf{x}'|^2}{2\theta^2}}$ is used, this leads to a closed-form kernel
\begin{equation}\label{curl_free_SE}
k_{\text{curl-free}}(\mathbf{x},\mathbf{x}') = 
\frac{\gamma^2}{\theta^2} 
e^{-\frac{|\mathbf{x} - \mathbf{x}'|^2}{2\theta^2}}
\left(
I_d - 
\left( \frac{\mathbf{x} - \mathbf{x}'}{\theta} \right)
\left( \frac{\mathbf{x} - \mathbf{x}'}{\theta} \right)^\top
\right); 
\end{equation}
see \citet{jidling2017linearly} or \citet{jidling2017strain}. We have derived this for dimension $d = 3$, but it is valid in any dimension $d \ge 2$ \citep{macedo2010learning}. The specific covariance kernel \eqref{curl_free_SE} was introduced by \citet{fuselier2007refined} in the context of reproducing kernel Hilbert spaces, and was also used by \citet{macedo2010learning,baldassarre2010vector,wahlstrom2013modeling,wahlstrom2015modeling,solin2018modeling}. 

In a similar way, one can enforce a divergence-free condition $\nabla \cdot \mathbf{f} = 0$ for a vector-valued GP $\mathbf{f}$ by writing $\mathbf{f} = \nabla \times \mathbf{g}$ and placing a GP prior on a vector field $\mathbf{g}$, as $\nabla \cdot (\nabla \times \mathbf{g}) = 0$ \citep{jidling2017linearly}.
Modeling the components of $\mathbf{g}$ as independent and placing a diagonal matrix-valued squared-exponential kernel on it leads to divergence-free covariance kernels for $\mathbf{f}$ of the form
\begin{equation}\label{div_free_SE}
k_{\text{div-free}}(\mathbf{x},\mathbf{x}') = 
\frac{\gamma^2}{\theta^2} 
e^{-\frac{|\mathbf{x} - \mathbf{x}'|^2}{2\theta^2}}
\left(
\left( \frac{\mathbf{x} - \mathbf{x}'}{\theta} \right)
\left( \frac{\mathbf{x} - \mathbf{x}'}{\theta} \right)^\top
+
\left(
(d-1) - \frac{\| \mathbf{x} - \mathbf{y} \|^2}{\theta^2}
\right) I_d
\right); 
\end{equation}
see \citet{macedo2010learning,wahlstrom2015modeling,jidling2017strain,baldassarre2010vector}. 
The specific divergence-free kernel \eqref{div_free_SE} appears to have been introduced by \citet{narcowich1994generalized}.

In all these examples, solving the key operator equation \eqref{operator_equation} has been an easy application of vector calculus identities. The work of \citet{jidling2017linearly} proposes an approach for solving \eqref{operator_equation} for general linear constraints involving first-order differential operators, which generalizes the curl-free and divergence-free examples. However, solving \eqref{operator_equation} in general is difficult, depends dramatically on $\mathcal{L}$, and may introduce significant computational challenges.
For example, to constrain a scalar function on the unit disk $D = \{ z = (x_1,x_2) \ : \ |z| < 1 \}$ in $\mathbb{R}^2$
to satisfy Poisson's equation
\begin{equation}
\Delta u = \frac{\partial^2 u}{\partial x_1^2} + \frac{\partial^2 u}{\partial x_2^2} = 0 \quad \text{ in } \quad D
\end{equation}
i.e., the constraint \eqref{transformed_constraint} with $\mathcal{L}_{\mathbf{x}} = \Delta$, one could exploit Poisson's kernel formula
\begin{equation}\label{poisson_formula}
u(r,\theta) = \mathcal{G} g = \frac{1}{2\pi} \int_{-\pi}^{\pi} P_r(\theta - t) g(e^{it}) dt 
\end{equation}
for a boundary value $g$ defined on $\partial D$. More precisely, $g \in L^1(\mathbb{T})$. Thus, one could model $g$ with an appropriate GP with covariance kernel $k_g$ and use $\mathcal{G}$ in \eqref{poisson_formula} to satisfy \eqref{operator_equation} and then to define a kernel $k_u$ via \eqref{transformed_kernel}. However, in this case $G$ is an integral operator which would make evaluation of \eqref{transformed_kernel} more difficult than the vector calculus examples discussed above. This illustrates that imposing the constraint via solving the operator equation \eqref{operator_equation} requires using analytical representations of the solution to the constraint equation \eqref{transformed_constraint} that vary significantly from case to case. The same issue is illustrated by an example of a biharmonic constraint in one-dimension for a scalar GP in \citet{jidling2017strain}. This is in contrast to the block covariance method of \ref{pde_constraints}, which involves more straightforward computation of the kernel blocks in \eqref{eq:joint_covariance}.  

\subsection{Empirical Mean and Covariance}
\label{sec:empirical}
Given an ensemble of realizations of a random field $Y$  on a set of grid points and a smaller set of high-fidelity data on a subset of the low-fidelity grid points, %this section presents the method of
\citet{yang} build a Gaussian process for the unknown field over the unstructured grid which also passes through the high-fidelity data, at the same time ensuring that the GP satisfies the PDE used to generate the low-fidelity ensemble. The ensemble data may be obtained from a large number of simulations over a single unstructured grid of a deterministic solver for a linear PDE, sampling the stochastic parameters in the PDE according to some distribution. The high-fidelity may consist of field data obtained through a costly experiment, a situation common in geostatistics.

%Another approach to physics-constrained Gaussian processes is to generate an ensemble of realizations of 
%a stochastic system, such as realizations of solutions to stochastic partial differential equations.  
%The ensemble represents a collection of approximations of the random field $Y$ that one is trying 
%to identify with the Gaussian process.  
The idea of the article of \citet{yang} is to compute the mean and covariance function of the GP empirically from these realizations of the random field $Y$.  This removes the need to infer the hyperparameters of a
covariance function.  Instead, one simply calculates the mean and covariance matrix 
from the random field realizations as follows.  Using the notation of \citet{yang}, we 
assume that we have $M$ realizations $Y^m(\mathbf{x})$ of the output field $Y(\mathbf{x})$ for $\mathbf{x}$ in the $d$-dimensional grid 
$\{\mathbf{x}_i\}_{i=1}^N$ (the low-fidelity data). Then the mean and covariance kernel are respectively given by
\begin{equation}
\mu(\mathbf{x}) \approx \mu_{\text{MC}}(\mathbf{x})=\frac{1}{M}\sum_{m=1}^{M}Y^m(\mathbf{x})
\end{equation} 
and
\begin{equation}
k(\mathbf{x},\mathbf{x}') \approx k_{\text{MC}}(\mathbf{x},\mathbf{x}')=\frac{1}{M-1}\sum_{m=1}^{M}{(Y^m(\mathbf{x})-\mu_{\text{MC}}(\mathbf{x}))(Y^m(\mathbf{x}')-\mu_{\text{MC}}(\mathbf{x}'))}.
\end{equation}
Hence, with $\mathbf{Y}^m = [Y^m(\mathbf{x}_1), ..., Y^m(\mathbf{x}_N)]^\top$ and $\bm{\mu}_{\text{MC}} = [\mu_{\text{MC}}(\mathbf{x}_1), ..., \mu_{\text{MC}}(\mathbf{x}_N)]^\top$, the covariance matrix is approximated by
\begin{equation}
\mathbf{C} \approx \mathbf{C}_{\text{MC}}=\frac{1}{M-1}\sum_{m=1}^{M}{(\mathbf{Y}^m-\boldsymbol{\mu}_{\text{MC}})(\mathbf{Y}^m-\boldsymbol{\mu}_{\text{MC}})^\top}.
\end{equation}
The above formulas for the mean and covariance of the GP over the unstructured grid $\{\mathbf{x}_i\}_{i=1}^N$ can then be used in the usual prediction formula \eqref{eq:gpreg} for the posterior mean and variance at any point in the grid, conditioned on high-fidelity data at a \emph{subset} of points on the grid. 

It is important to note that this approach does not assume stationarity of the GP, nor does it 
assume a specific form of the covariance function. \citet{yang} have shown that physical constraints in the form of a deterministic linear operator are guaranteed to be satisfied within a certain error in the resulting prediction when using this approach. The method was extended to model discrepancy between the low- and high-fidelity data in \citet{yang2019physics}. They also provide an estimate of the error in preserving the physical constraints. However, as the method uses an empirical mean and covariance, it \emph{cannot} interpolate for the field between the points where the stochastic realizations are available. The step of GPR for prediction at an arbitrary point $\mathbf{x}^*$, represented by \eqref{eq:gpreg}, is not available, as the covariance kernel function is bypassed entirely;
%In other words, there is no formula for the covariance that can be evaluated at $\mathbf{x}^*$ if realizations of the stochastic field $Y$ were not available at $\mathbf{x}^*$; 
the covariance is obtained directly in matrix form over the unstructured grid.

\REVISION{
\subsection{Specialized Kernel Construction}
\label{sec:expansion}
\citet{Albert} developed an approach that is suited for GPR with linear PDE constraints of the form \eqref{eq:pde_constraint} for the case of vanishing or localized source terms $f$. In the latter case, the solution $u$ is represented as a solution to the homogeneous equation $\mathcal{L}u = f$ and an inhomogeneous contribution obtained as a linear model over fundamental solutions corresponding to the point sources.

Focusing on the GPR for the solution $u$ to the homogeneous equation $\mathcal{L} u = 0$, a specalized kernel function $k$ is derived from $\mathcal{L}$ such that the GPR prediction satisfies the equation exactly. In this sense, the approach is similar to that of Section \ref{transformed_covariance}, although the kernel is not obtained from a transformation of a prior kernel but rather is constructed from solutions to the problem $\mathcal{L}u = 0$. \citet{Albert} show that such a covariance kernel must satisfy
\begin{equation}\label{eq:albert_condition}
\mathcal{L}_{\mathbf{x}} k(\mathbf{x}, \mathbf{x}')\mathcal{L}^\top_{\mathbf{x}'} = 0
\end{equation}
in the notation of Section \ref{transformed_covariance}, and seek kernels $k$ in the form of a Mercer series
\begin{equation}\label{eq:albert_kernel}
k(\mathbf{x}, \mathbf{x}') = \sum_{i,j} \phi_i(\mathbf{x}) \Sigma^{i,j}_p \phi_j(\mathbf{x}')
\end{equation}
for basis functions $\phi_i$ and matrix $\Sigma_p$. They point out that convolution kernels can also be considered. \citet{Albert} study the Laplace, heat, and Hemholtz equations, performing MLE and inferring the solution $u$ from the PDE constraint and scattered observations. They construct kernels of the form \eqref{eq:albert_kernel} which satisfy \eqref{eq:albert_condition} by selecting $\{\phi_i\}$ to be an orthogonal basis of solution to the corresponding equations. Although the resulting kernels are not stationary and require analytical construction, they result in improved reconstructions of solutions from the observations compared to squared-exponential kernels. We note that similar constructions of kernels -- as expansions in a suitable basis -- are utilized in the approach of \cite{solin2019know} in the following section to enforce boundary conditions. 
}

\section{Boundary Condition Constraints}\label{sec:boundary_constraints}
Boundary conditions and values are yet another type of prior knowledge that may be incorporated into Gaussian process regression.   
In many experimental setups, measurements can be taken at the boundaries of a system in a cheap and non-invasive way that permits nearly complete knowledge of the boundary values of an unknown field. In other cases, the boundary values may be fully known or controlled by the user, such as for a system in a heat bath. Theoretically, various boundary conditions are often needed to complete the description of a well-posed model. Thus, intrinsic boundary condition constraints on a GP, as opposed to the treatment of boundary measurements as scattered data, may be of interest in applications both for improved accuracy and to avoid the computational burden of an expanded dataset. For one-dimensional GPs, the enforcing Dirichlet boundary conditions is trivial; noiseless observations at the boundary can be used to produce a posterior mean and covariance that satisfy the boundary conditions exactly. In higher dimensions, however, it is nontrivial to constrain GPs to satisfy boundary conditions globally over a continuous boundary. \REVISION{\citet{graepel2003solving} constructed an example of GPR on the two-dimensional unit square $[0,1]^2$ with Dirichlet boundary conditions by writing the solution as a product of a factor represented by a GP and an analytic factor which was identically zero at the boundary. We discuss a more general approach based on spectral expansions below.}

\subsection{Spectral Expansion Approach}
\label{spectral_expansion_approach}

The work of \citet{solin2019know} introduced a method based on the spectral expansion of a desired stationary isotropic covariance kernel 
\begin{equation}\label{e:stationary_isotropic_kernel}
k(\mathbf{x},\mathbf{x}') = k(|\mathbf{x}-\mathbf{x}'|)
\end{equation}
in eigenfunctions of the Laplacian. For enforcing zero Dirichlet boundary values on a domain $\Omega$, \citet{solin2019know} use the \emph{spectral density} (Fourier transform) of the kernel \eqref{e:stationary_isotropic_kernel},
\begin{equation}
\label{e:spectral_density}
s(\bm{\omega}) = \int_{\mathbb{R}^d}
e^{-i \bm{\omega} \cdot \mathbf{x}} 
k(|\mathbf{x}|)
d\mathbf{x}. 
\end{equation}
This enters into the approximation of the kernel:
\begin{equation}\label{e:spectral_expansion_kernel}
k(\mathbf{x},\mathbf{x}') \approx \sum_{\ell=1}^m 
s(\lambda_\ell) \phi_\ell(\mathbf{x}) \phi_\ell(\mathbf{x}'),
\end{equation}
where $\lambda_j$ and $\phi_j$ are the Dirichlet eigenvalues and eigenfunctions, respectively, of the Laplacian on the domain $\Omega$.
In \eqref{e:spectral_expansion_kernel}, $s(\cdot)$ is thought of as a function of a scalar variable; since $k$ is isotropic in \eqref{e:stationary_isotropic_kernel}, so is the Fourier transform $s(\bm{\omega}) = s(|\bm{\omega}|)$.  
Note that the expansion \eqref{e:spectral_expansion_kernel} yields a covariance that is zero when $\mathbf{x} \in \partial\Omega$ or $\mathbf{x}' \in \partial\Omega$.  
Thus if the mean of the GP satisfies the zero boundary conditions, Gaussian process predictions using the series \eqref{e:spectral_expansion_kernel} will satisfy the boundary condition as well. 

\subsection{Implementation}
The first implementation task that presents itself is computation of the Dirichlet spectrum $(\lambda_\ell, \phi_\ell)$ of the Laplacian
\begin{align}
%\begin{cases}
\Delta \phi_\ell &= \lambda_\ell \phi_\ell \quad\text{ in } \Omega \\
\phi_\ell &= 0 \quad\text{ on } \partial\Omega
%\end{cases}
\end{align}
For basic domains, such as rectangles, cylinders, or spheres, this can be solved in closed form. For general domains, the problem must be discretized and an approximate spectrum computed. 
\citet{solin2019know} obtain an approximate spectrum by discretizing the Laplace operator with a finite difference formula and applying a correction factor to the eigenvalues of the resulting matrix. There are many other approaches for computing the spectrum of the Laplacian with various boundary conditions; see, e.g., \citet{song2017computing} for an approach using the spectral element method for calculating both Dirichlet and Neumann spectrum in complex geometries. 
Evaluation of $s(\lambda_\ell)$, where $s$ denotes the spectral density \eqref{e:spectral_density} in \eqref{e:spectral_expansion_kernel}, is typically not difficult since $s$ is available in closed form for many stationary kernels, such as the squared exponential (SE) and Mat\'ern ($M_{\nu}$) kernels:
\begin{align}
s_{\text{SE}}(|\bm{\omega}|; \gamma, \theta) &=
\gamma^2 (2\pi\theta^2)^{\frac{d}{2}}e^{-\frac{|\bm{\omega}|^2 \theta^2}{2}},\\
s_{M_{\nu}}(|\bm{\omega}|; \gamma, \theta) &= 
\gamma^2 \frac{2^d \pi^{\frac{d}{2}} (2\nu)^\nu \Gamma(\nu + \frac{d}{2})}
{\theta^{2\nu}\Gamma(\nu) }
\left(
\frac{2\nu}{\ell^2} + |\bm{\omega}|^2
\right)^{-\frac{2\nu + d}{2}}. 
\end{align}

Next, we review from \citet{solin2019know} and \citet{solin2019hilbert} how the formulas for Gaussian processes regression and training can be expressed using the the formulation \eqref{e:spectral_expansion_kernel}. 
Given $n$ data points $\{(\mathbf{x}_i, y_i)\}_{i=1}^n$, 
the covariance matrix is approximated
using \eqref{e:spectral_expansion_kernel} as
\begin{equation}
K_{ij} = k(\mathbf{x}_i,\mathbf{x}_j) \approx 
\sum_{\ell=1}^m 
\phi_{\ell}(\mathbf{x}_i) s(\lambda_\ell) 
\phi_{\ell}(\mathbf{x}_j).
\end{equation}
Introducing the $n \times m$ matrix $\Phi$,
\begin{equation}
{\Phi}_{i \ell} = 
\phi_{\ell}(\mathbf{x}_i), 
\quad 1 \le i \le n,
\quad 1 \le \ell \le m, 
\end{equation}
and the $m \times m$ matrix 
%\begin{equation}\label{e:def_lambda_bc}
$\Lambda = \text{diag}(s(\lambda_{\ell})), 1 \le \ell \le m,$
%\end{equation}
this can be written
\begin{equation}\label{e:big_k_spectral_matrix}
{K} \approx {\Phi} \Lambda {\Phi}^\top.  
\end{equation}
Thus, the covariance matrix $K$ is diagonalized and, for a point $\mathbf{x}^*$, we can write the $n \times 1$ vector
\begin{equation}\label{e:little_k_spectral_matrix}
\mathbf{k}_* = \left[k(\mathbf{x}^*, \mathbf{x}_i)\right]_{i=1}^n \approx 
\left[
\sum_{\ell=1}^m 
\phi_{\ell}(\mathbf{x}_i) s(\lambda_\ell) 
\phi_{\ell}(\mathbf{x}^*)
\right]_{i=1}^n
=
\Phi \Lambda \bm{\Phi}_*, 
\end{equation}
where the $m \times 1$ vector $\bm{\Phi}_*$ is defined by
\begin{equation}
\left[\bm{\Phi}_*\right]_{\ell} = \phi_\ell(\mathbf{x}^*),
\quad
1 \le \ell \le m.
\end{equation}
The Woodbury formula can be used to
obtain the following expressions for the posterior mean and variance over
a point $\mathbf{x}^*$ given a Gaussian likelihood 
$y_i = f(x_i)+\epsilon_i, \epsilon_i \sim \mathcal{N}(0,\sigma^2)$ \citep{solin2019know}:
\begin{align}\label{e:regression_formulas_bc}
\begin{split}
\mathbb{E}[f(\mathbf{x}^*)] &= \mathbf{k}_*^\top 
(K + \sigma^2 I)^{-1} \mathbf{y} \\
&=
\bm{\Phi}_*^\top 
(\Phi^\top \Phi + \sigma^2 \Lambda^{-1} )^{-1}
\Phi^\top \mathbf{y}. \\
\mathbb{V}[f(\mathbf{x}^*)] &= 
k(\mathbf{x}^*,\mathbf{x}^*) - 
\mathbf{k}_*^\top (K + \sigma^2 I)^{-1} \mathbf{k}_* \\
&=
\sigma^2 \bm{\Phi}_*^\top 
(\Phi^\top \Phi + \sigma^2 \Lambda^{-1})^{-1} \bm{\Phi}_*.
\end{split}
\end{align}
Strategies for using this method with non-Gaussian likelihoods are also discussed by \citet{solin2019know}, although we do not go over them here. 
For use in hyperparameter training, the following formulas were derived in \citet{solin2019hilbert} and \citet{solin2019know} for the 
negative log-marginal-likelihood 
\begin{align}
\label{e:nlml_bc}
&\phantom{\partial}\begin{multlined} 
-p(\bold{y}|X,\bm{\theta}) =
\frac{n-m}{2} \log \sigma^2
+\frac{1}{2} \sum_{\ell = 1}^{m} 
{\color{black}\log\Big(}
\Lambda_{\ell,\ell}
{\color{black}\Big)}
+\frac{1}{2} \log \det \left( \sigma^2 \Lambda^{-1} + \Phi^\top \Phi \right)
+\frac{n}{2} \log(2\pi) \\
+\frac{1}{2\sigma^2}\left[
\mathbf{y}^\top \mathbf{y} - \mathbf{y}^\top \Phi
\left(\sigma^2 \Lambda^{-1} + \Phi^\top \Phi \right)^{-1} \Phi^{\color{black}\top}
 \mathbf{y}
\right],
\end{multlined}
\end{align}
and in \citet{solin2019hilbert} for its derivative:
\begin{align}
\label{e:nlml_bc_derivatives}
\begin{split}
&\begin{multlined}
-\frac{\partial p(\bold{y}|X,\bm{\theta})}{\partial \theta_k} = 
\frac{1}{2} \sum_{\ell = 1}^m \frac{1}{\Lambda_{\ell,\ell}} 
\frac{\partial \Lambda_{\ell,\ell}}{\partial \theta_k}
-
\frac{\sigma^2}{2} \text{Tr}\left(
\left(\sigma^2 \Lambda^{-1} + \Phi^\top \Phi\right)^{-1}
\Lambda^{-2} \frac{\partial \Lambda}{\partial \theta_k}
\right) \\
-\mathbf{y}^\top \Phi 
\left( \sigma^2 \Lambda^{-1} + \Phi^\top \Phi \right)^{-1}
\left( \Lambda^{-2} \frac{\partial \Lambda}{\partial \theta_k} \right)
\left( \sigma^2 \Lambda^{-1} + \Phi^\top \Phi \right)^{-1}
\Phi^\top \mathbf{y}
,\end{multlined}
\\
&\begin{multlined}
-\frac{\partial p(\bold{y}|X,\bm{\theta})}{\partial \sigma^2} = 
\frac{n-m}{2\sigma^2}
+
\frac{1}{2}
\text{Tr}
\left(
\left( \sigma^2 \Lambda^{-1} + \Phi^\top \Phi \right)^{-1}
\Lambda^{-1}
\right)
\\ + \frac{1}{2\sigma^2}
\mathbf{y}^\top \Phi \left( \sigma \Lambda^{-1} + \Phi^\top \Phi \right)^{-1}
\Lambda^{-1}
\left( \sigma \Lambda^{-1} + \Phi^\top \Phi \right)^{-1}
\Phi^\top \mathbf{y} \\
\REVISION{-
\frac{1}{2\sigma^4} \left[
\mathbf{y}^\top \mathbf{y} - \mathbf{y}^\top \Phi
\left(\sigma^2 \Lambda^{-1} + \Phi^\top \Phi \right)^{-1} \Phi^{\color{black}\top}
 \mathbf{y}
\right].}
\end{multlined}
\end{split}
\end{align}
Note that $\Lambda$ is defined by the spectral density $s$ of the kernel $k$, which clearly depends on the kernel hyperparameters $\bm{\theta} = [\theta_i]$, however $\Phi$ does not. Typically, derivatives of $\Lambda$ with respect to $\theta_i$ can be computed in closed form which along with the formulas \eqref{e:nlml_bc} and \eqref{e:nlml_bc_derivatives} enable accurate first-order optimization of the kernel hyperparameters. 

\subsection{Extensions}
The expansion \citet{solin2019hilbert} was originally developed for the computational advantages of using a low rank approximation to a kernel (see Section \ref{sec:spectral_low_rank} for a discussion of this aspect) rather than for boundary condition constraints. Consequently, the discussions in \citet{solin2019know} and \citet{solin2019hilbert} focused only on periodic and zero Dirichlet boundary conditions. One possible way 
to constrain a Gaussian process $f$ to satisfy nonzero Dirichlet conditions would be to write $f = (f - g) + g$, where $g$ is a harmonic function that satisfies a given nonzero Dirichlet condition, and model $f-g$ as a Gaussian processes that satisfying a zero Dirichlet condition using the above approach. \citet{solin2019know} remark that the method could also be extended to Neumann boundary conditions by using the Neumann eigenfunctions of the Laplacian, although no examples are given. Another limitation is that spectral expansions in \citet{solin2019know} and \citet{solin2019hilbert} are only considered for isotropic kernels, but they suggest that the approach can be extended to the nonisotropic case.

\section{Computational Considerations}\label{sec:computation_considerations}
In this section, we describe methods that can help reduce the computational cost of constructing constrained GP models. Typically, building a constrained GP is significantly more expensive than training an unconstrained GP because of larger data sets representing derivative constraints, bounds, etc. at virtual points. Consequently, computationally efficient strategies for building constrained GPs are paramount.
In Section \ref{sec:mvn} we discuss the truncated multivariate normal distribution, which is a fundamental component of the approaches discussed in Sections \ref{sec:transform_output},\ref{sec:daveiga},\ref{sec:splines} and \ref{daveiga_monotonic}. We then discuss the related problem of maximum likelihood estimation of the hyperparameters of constrained GPs constructed using the spline approach discussed in Sections \ref{sec:splines}, \ref{sec:splines_monotonic}, and \ref{sec:convexity}. 
The final subsection \ref{sec:low_rank} focuses on reducing the numerical linear algebra cost of inference, using low-rank and Kronecker methods, respectively. The majority of approaches surveyed in these two sections were developed for unconstrained GPs; however, some methods have been applied in the constrained setting. Since such numerical recipes are the focus of much deeper survey articles such as \citet{quinonero2005RR} and \citet{quinonero2005JMLR}, we have intentionally kept our discussion short, while providing references to applications in constrained GPR where available.

\subsection{The Truncated Multivariate Normal Distribution}
\label{sec:mvn}

Given a positive-definite covariance matrix $\Sigma$ and \REVISION{vectors $\mathbf{a},\mathbf{b} \in \mathbb{R}^d$ defining a rectangle 
$\{\mathbf{a} \le \mathbf{x} \le \mathbf{b}\}$}, the truncated normal distribution is the conditional distribution of the random variable $\bold{x} \sim \mathcal{N}\left(\bm{\mu},  \Sigma\right)$ given \REVISION{$\mathbf{a} \le \mathbf{x} \le \mathbf{b}$}. The \REVISION{density $\mathcal{TN}\left(\bm{\mu},  \Sigma, \mathbf{a}, \mathbf{b} \right)$} of the truncated normal can be expressed as
\begin{equation}\label{e:truncated_normal}
\REVISION{\mathcal{TN}\left(\bold{x}; \bm{\mu},  \Sigma, \mathbf{a}, \mathbf{b} \right)
=
\frac{\mathds{1}_{\{\mathbf{a} \le \mathbf{x} \le \mathbf{b}\}}(\bold{x})}{C}
\mathcal{N}\left(\bold{x}; \bm{\mu},  \Sigma \right),}
\end{equation}
where the normalization constant
\begin{align}\label{e:truncated_normalization}
\begin{split}
C
&=
\int_{a_1}^{b_1}
\int_{a_2}^{b_2}
...
\int_{a_d}^{b_d}
\mathcal{N}\left(\mathbf{x}; \bm{\mu},  \Sigma\right)
d\mathbf{x}_1
d\mathbf{x}_2
...
d\mathbf{x}_d\\
&=
\frac{1}{(2\pi)^{\frac{d}{2}}|\Sigma|^{\frac{1}{2}}}
\int_{a_1}^{b_1}
\int_{a_2}^{b_2}
\hdots
\int_{a_d}^{b_d}
\exp\left({-\frac{1}{2} (\mathbf{x} - \bm{\mu})^\top \Sigma^{-1} (\mathbf{x} - \bm{\mu})}\right)
d\mathbf{x}_1
d\mathbf{x}_2
...
d\mathbf{x}_d
\end{split}
\end{align}
is the probability that a sample of $\mathcal{N}(\REVISION{\bm{\mu}}, \Sigma)$ lies in \REVISION{$\{\mathbf{a} \le \mathbf{x} \le \mathbf{b}\}$}. %The set $S$ can be any Lebesgue measurable set, but in this survey we consider it to be a rectangle or more generally a \REVISION{polytope} $\mathcal{C}$ defined by a finite set of linear inequalities, as in Section \ref{sec:splines}. 

For general $\Sigma$ and dimension $d$, computing the normalization constant and sampling from the truncated multinormal distribution 
\eqref{e:truncated_normal} can be difficult and require specialized methods. Of course, from the definition \eqref{e:truncated_normalization} these two problems are related. However, they appear in two different contexts. Calculating \REVISION{integrals of the form \eqref{e:truncated_normalization}, known as \emph{Gaussian orthant probabilities},} is called for in constrained maximum likelihood estimation of the GPR hyperparameters, while sampling \eqref{e:truncated_normal} is needed for posterior prediction in several approaches discussed above. Therefore, we discuss sampling first, and discuss evaluation of \REVISION{Gaussian orthant probabilities} in the next Section \ref{sec:mle}. 

While there are several possible approaches to sampling from \eqref{e:truncated_normal}, simple Monte Carlo methods scale poorly to high dimensions. One such example -- rejection sampling from the mode -- was discussed in Section \eqref{sec:spline_sampling}. 
In principal, it is possible to use a Metropolis-Hastings approach to sample the values of the knots, but it is expected that the dimensionality of the chain for a large number of splines is likely to slow down the convergence of the chain.
Several Markov Chain Monte Carlo (MCMC) methods were studied by \citet{lopez2018} for sampling the truncated multivariate normal posterior distribution that arises in the spline approach described in Section \ref{sec:splines}. 
Comparison of expected sample size metrics suggested that Hamiltonian Monte Carlo (HMC) is the most efficient sampler in the setting of that article.  An different approach for sampling \eqref{e:truncated_normal}, based upon elliptical slice sampling and the fast Fourier transform, was presented in \citet{ray_pati_bhattacharya}.

\subsection{Constrained Maximum Likelihood Estimation for Splines}\label{sec:mle}
We review the work of \citet{lopez2018} which discusses maximum likelihood estimation of hyperparameters within the spline approach \REVISION{discussed in Sections \ref{sec:splines} and \ref{sec:splines_monotonic}}. 
The starting point is the constrained log-marginal-likelihood function given the constraints $\bm{\xi} \in \mathcal{C}$, \REVISION{where we have denoted $\mathcal{C} = \{\mathbf{a} \le \bm{\xi} \le \mathbf{b}\}$}.
This is based on the posterior \REVISION{density} $p_{\bm{\theta}}(\mathbf{y} | \bm{\xi} \in \mathcal{C})$ of $\mathbf{y}$ given the constraint $\bm{\xi} \in \mathcal{C}$, which by Bayes' rule can be expressed as
\begin{equation}
p_{\bm{\theta}}(\mathbf{y} | \bm{\xi} \in \mathcal{C}) = 
\frac{p_{\bm{\theta}}(\mathbf{y}) P_{\bm{\theta}}(\bm{\xi} \in \mathcal{C} | \Phi \bm{\xi} = \mathbf{y})}{P_{\bm{\theta}}(\bm{\xi} \in \mathcal{C})}.
\end{equation}
Taking the logarithm yields a constrained log-marginal-likelihood function:
\begin{align}
\begin{split}
\label{e:cmle}
\mathcal{L}_{\text{cMLE}}
&=
\log p_{\bm{\theta}}(\mathbf{y} | \bm{\xi} \in \mathcal{C}) \\
&= 
\log p_{\bm{\theta}}(\mathbf{y}) + \log P_{\bm{\theta}}(\bm{\xi} \in \mathcal{C} | \Phi \bm{\xi} = \mathbf{y})  - \log P_{\bm{\theta}}(\bm{\xi} \in \mathcal{C}) \\
&=\mathcal{L}_{\text{MLE}} + \log P_{\bm{\theta}}(\bm{\xi} \in \mathcal{C} | \Phi \bm{\xi} = \mathbf{y})  - \log P_{\bm{\theta}}(\bm{\xi} \in \mathcal{C}).
\end{split}
\end{align}
In the first term, $p_{\bm{\theta}}(\mathbf{y})$ refers to the probability density function of the random variable $\mathbf{y}$ with hyperparameters $\bm{\theta}$; thus, the first term is simply the unconstrained log-marginal-likelihood \eqref{eq:log_like} which we denote $\mathcal{L}_{\text{MLE}}$. In the second and third terms, $P_{\bm{\theta}}$ refers to the probability of the indicated events. 
As $\bm{\xi}$ and  $\bm{\xi} | \{\Phi \bm{\xi} = \mathbf{y}\}$ are both normally distributed by equations \eqref{eq:multivar_norm} and \eqref{eq:gpreg}, respectively, the two last terms in \eqref{e:cmle} can be expressed as integrals of a normal \REVISION{density} over 
$\mathcal{C}$, just like the normalization constant \eqref{e:truncated_normalization}.
\REVISION{Such integrals can be reduced to integrals over orthants, so the last two terms in \eqref{e:cmle} are referred in \citet{lopez2018} as Gaussian orthant probabilities.}

Unlike the sampling of \eqref{e:truncated_normal}, for which computing such integrals can be avoided with MCMC, calculation of Gaussian orthant probabilities is unavoidable if the user wants to train the kernel hyperparameters using the constrained objective function \eqref{e:cmle}, which we refer to as cMLE. A thorough discussion of numerical approaches to truncated Gaussian integrals is \citet{genz2009computation}. \citet{lopez2018} utilize the minimax exponential tilting method of \citet{botev2017normal}, reported to be feasible for quadrature of Gaussian integrals in dimensions as high as 100, to compute the Gaussian orthant probabilities in \eqref{e:cmle} and compare cMLE with MLE. Another current drawback of cMLE is that the gradient of $\mathcal{L}_{\text{cMLE}}$ is not available in closed form, unlike the gradient of 
$\mathcal{L}_{\text{MLE}}$ \citep{rasmussen}. 
Thus, in \citet{lopez2018}, MLE was performed using a L-BFGS optimizer, while cMLE was performed using the method of moving asymptotes. This involved a numerical approximation to the gradient of $\mathcal{L}_{\text{cMLE}}$, which in our experience can impact the accuracy of the optimization. Although these numerical differences hamper direct comparison of MLE and cMLE, it was found by \citet{lopez2018} that for the case of limited data, cMLE can provide more accurate estimation of hyperparameter values and confidence intervals than MLE. 

\citet{lopez2018} also studied under which conditions MLE and cMLE yield consistent predictions of certain hyperparameters. This was further studied in \citet{bachoc2019}, in which the authors perform an analysis of MLE and cMLE for the case of fixed-domain asymptotics, i.e., data in a fixed domain, as the number of data points tends to infinity. In this regime of dense data, the effect of constraints is expected to diminish. The authors show that MLE and cMLE yield consistent hyperparameters in this limit for the case of boundedness, monotonicity, and convexity constraints, and suggest quantitative tests to determine if the number of data points is sufficient to suggest unconstrained MLE as opposed to the more expensive cMLE.

\subsection{Scalable Inference}
\label{sec:low_rank}

%The standard approach for Gaussian process regression uses the entire training dataset to infer the function value at a new point, but as described in Section \ref{sec:gpr}, this inference is expensive and scales poorly with large datasets; covariance matrix inversion scales as $O(N^3)$. This scaling motivates the use of lower rank representations of the covariance or full dataset to reduce this cost \citep{quinonero2005RR}.  
As pointed out in Section \ref{sec:gpr}, inference in GPR using the entire training dataset (of size $N$) scales as $O(N^3)$ due to covariance matrix inversion. This is exacerbated by certain methods to enforce constraints, such as the linear PDE constraints in Section \ref{pde_constraints}, which require the inclusion of ``virtual'' constraint points in the training data.There have been few studies on improving scalability of constrained GPs. Thus, in this section, we mention several promising approaches and possible applications to constrained GPs. 
%The low rank methods for matrix inversion discussed below can be placed into two categories. The first category consists of methods that exploit properties of the algorithms used to assemble the GP covariance matrix. These methods, 
Some strategies, including the subset of data approach, the inducing point approach, and the spectral expansion approach, are specific to covariance matrices of GPs. Other methods are based on general linear algebra techniques. 

\subsubsection{Subset of data \& Inducing point methods}
\label{sec:subset_of_data}

One notable feature of increasing the density of training data is that the covariance matrix tends to become more ill-conditioned, the result of partially redundant information being added to the matrix.
%; if a cluster of points near each other are added to the covariance matrix, then the cost of the inference is increased without improving the quality of the predictions. In such situations it is worthwhile to identify an optimal subset of the data that may be used to perform inference, i.e.
In such situations it is worthwhile to identify a \emph{subset of data} that minimizes prediction error subject to a maximum dataset size constraint \citep{quinonero2005JMLR}. 
%Due to the combinatorial complexity of identifying such an optimal set of points, it is more common to either use greedy methods to select a subset or use multiple randomly selected sets of points to form individual GPs and combine the predictions.
%Greedy methods randomly select a subset of points to form a GP, and then add points one-at-a-time to the initial set until the computational budget has been consumed. At this point, the hyperparameters for the GP are optimized and the process repeated with a new subset of points and the updated hyperparameters until a satisfactorily converged GP has been achieved.
%Successfully applying this method requires a procedure to judiciously select new points to add to the current subset.
Greedy methods involve sequentially choosing points in the domain that have the maximal predictive variance in order to reduce the uncertainty in the final GP.
This choice is natural and has connections to information-theoretic metrics;
%and tends to choose points that are maximally separated from each other.
other metrics include cross-validation prediction error, the likelihood value, or Bayesian mean-square prediction error. 
%The steps used to update the current GP with new points also affects the efficiency of this algorithm. 
Rather than building a new covariance matrix and inverting it for each added point, one may take advantage of the Woodbury matrix inversion lemma and block-matrix inversions to efficiently compute the inverse of the covariance matrix\citep{quinonero2005JMLR}.

Other methods for performing subset selection are based on \emph{local approximation}. 
%In general, and in particular for non-stationary processes, it may be desirable to make a prediction using only the data points in the vicinity of a query point. 
Frequently, the function values far away from a point of interest may have little influence on the function value there.
% and the covariance function that best describes the process locally may also change drastically at different points. 
%suggesting
%it may make more sense to target 
%regression that uses a subset of points from only the local region to better inform the Gaussian process. 
A simple strategy based on this idea is to select the nearest neighbors to the target point to form the prediction.
%, which is fast but may be suboptimal.
The local approximation GP~\citep{gramacy_lagp,gramacy_lagpR} approach combines such local approximation with a greedy search heuristic to identify a better set of points to minimize the mean-squared prediction error at the location of interest.

%The utility of using the subset of data approach is its simplicity and flexibility. The inference process with or without constraints does not change with this method and does not rely on any particular structure in the data. Any of the aforementioned approaches may be readily combined with bound constraints or monotonicity constraints with no added difficulty. 

%\subsubsection{Inducing point approximations}
%\label{sec:inducing_point}

Using a subset of points to form the GP corresponds to selecting a subset of the rows/columns of a full covariance matrix to represent the dataset. \citet{quinonero2005RR} generalize this to a broad set of low-rank approximations to the full covariance matrix based on \emph{inducing points}. In these methods, a subset (size $m$) of the data is used to form an approximate likelihood or prior for the entire dataset; all of the data is used, but most of the data is modeled as being conditionally dependent on a few inducing points. This reduces the cost of inference from $O(N^3)$ to $O(Nm^2)$. The greedy methods discussed above may be applied to identify an optimal set of inducing points. 

Such methods may be especially helpful for selection and placement of virtual points for enforcing constraints. However, to our knowledge, there have not been any studies of this. An important question is how to treat the two sets of data -- training and virtual -- using these approaches.

\subsubsection{Spectral expansion approximation}
\label{sec:spectral_low_rank}
%Here, we describe how the spectral expansion of the covariance kernel in $m$ basis functions used to enforce boundary conditions in Section \ref{spectral_expansion_approach} yields a low rank approximation to a general Gaussian process with covariance kernel $k(\bold{x},\bold{x}')$.  
%In that section, we followed the presentation of \cite{solin2019know}. In fact, the method was originally developed in \cite{solin2019hilbert} (see also \cite{hensman2017}) for the goal of reduced rank GPR, at the cost of introducing artificial boundary conditions.
%Here, the viewpoint is different to that of 
The approach of Section \ref{spectral_expansion_approach} for boundary condition constraints can also be used for reduced rank GPR \citep{solin2019hilbert,hensman2017}.
An expansion of a covariance kernel in terms of the eigenvalues of the Laplacian with periodic boundary values in an artificial box containing the data is used to approximate the covariance kernel, as in \eqref{e:spectral_expansion_kernel}. The error of approximation should be small if the boundaries of the box are sufficiently far from the data locations. 
With $m$ basis functions in the expansion \eqref{e:spectral_expansion_kernel}, the formula \eqref{e:regression_formulas_bc} implies that inverses are required only of matrices of size $m$. Therefore, inversion scales as $O(m^3)$, while multiplication for inference in \eqref{e:regression_formulas_bc} scales as $O(m^2 N)$. 
%Thus, the method only scales linearly with $n$, albeit requiring a high one-time cost of computing the spectrum $(\lambda_\ell, \phi_\ell)_{i=1}^{m}$ which may scale as high as $O(m^3)$.
Moreover, formulas \eqref{e:nlml_bc} and \eqref{e:nlml_bc_derivatives} and the fact that $\Phi$ does not depend on the kernel hyperparameters imply that the same reduced-rank advantage is present in hyperparameter training via MLE.
%More specific computational complexity and implementation details are discussed in \citet{solin2019hilbert} and \citet{solin2019know}, but it is clear when 
%$n \gg m$, the advantage may be considerable. 
%\citet{solin2019hilbert} apply this method to data from the airline industry with $N = O(10^6)$ samples in $8$ dimensions using $m = 8 \times 40$ basis functions. 

\subsubsection{Linear Algebra Techniques}
\label{sec:hierarchical_matrices}

%SVD
Rather than trying to reduce the effective size of the training and virtual constraint points, it is possible to simply approximate covariance matrix inversion using more general numerical linear algebra techniques.
%Besides the need to overcome the scaling of the inversion, the ill-conditioning of the covariance matrix may require alternative methods. Using low-rank approximations is a common approach. For example, 
We expect such methods to extend more readily to constrained GPs than the methods of Section \ref{sec:subset_of_data}, although they may be less optimal and may not inform the placement of virtual constraint points. 

The pseudo-inverse is often used to avoid the small eigenvalues that can corrupt predictions \citep{brunton2019data}, although the singular value decomposition or eigenvalue decomposition are both computationally expensive as well.
%hierarchical matrices
Hierarchical matrices are an efficient way of approximating the full covariance matrix in a manner amenable to fast inversion \citep{pouransari2017fast}. Sub-blocks of the matrix are replaced with fixed-rank approximations using a computational tree to organize the hierarchy of the matrix, and operations on the full matrix such as multiplication or inversion can be accomplished efficiently by operating on each of the individual ``leaves'' of the tree. \citet{geoga2019scalable} applied hierarchical matrices methods to maximum likelihood estimation for Gaussian processes.

%\subsubsection{Sparse linear algebra}
%\label{sec:sparse_linear_algebra}

%CLIME 
An alternative to inverting the covariance matrix is to
%attempt to compute
%solve for the inverse covariance matrix, referred to as the precision matrix, using direct methods \citep{zhao2013sparse}. 
%While not technically low-rank, these methods do approximate the traditional GP with a structure that embeds sparsity in the conditional dependence of the data.
%From a purely linear algebraic perspective, one may attempt to 
%That is, 
to set up an optimization problem for a matrix such that the error between the product of the matrix with the covariance matrix and the identity matrix is minimized \citep{zhao2013sparse}. The solution to this linear program is, of course, the precision matrix,
%, and it may be computed exactly at great expense and will be fully dense for a general covariance function. However, 
but by adding a $L_1$ penalty term on the entries of the matrix to the objective function as in LASSO regression, sparsity will be induced in the result. This estimator is referred to as CLIME \citep{cai2011}.

%GMRF
Another popular approach for modeling sparsity in random processes is to use a Gaussian Markov random field (GMRF) \citep{sorbye2014scaling, rue2002fitting}. In a GMRF, the data may be seen as forming an undirected graph where points close to each other in data-space are connected by an edge, and points far from each other are not. Thus while all points are correlated with each other, most are only conditionally dependent on each other; this translates to a sparse precision matrix where the only off-diagonal entries correspond to points that are connected to each other. Different choices of the covariance widths or kernels, such as truncated or tapered kernels \citep{shaby2012tapered,kaufman2008covariance}, yield different levels of sparsity in the final precision matrix.
%Perhaps the simplest approach is to use a truncated or tapered covariance function \citep{shaby2012tapered,kaufman2008covariance}. Depending on the magnitude of the covariance values or the tapering length, varying levels of sparsity may be induced in the covariance matrix, which can make linear algebra routines for the inversion process (Cholesky decomposition or matrix-vector multiplications) much faster.

\subsubsection{Hierarchical Decomposition for non-Gaussian likelihoods}
\label{kronecker}
%This section describes the approach of \citet{flaxman2015} for scalable Kronecker methods for inference in
%GPs with non-Gaussian likelihoods on dense multi-dimensional grids. 
Constraints that rely on non-Gaussian likelihood were reviewed in Sections \ref{sec:likelihood_formulations} and \ref{constrained_likelihood_with_derivative_information}. Recent work of \citet{flaxman2015} focuses on scalable inference with non-Gaussian likelihoods on dense multi-dimensional grids.
The key assumption enabling the use of Kronecker formulas is that 
the inputs $X$ are on a multi-dimensional Cartesian grid
\begin{equation}
X=X_1\otimes X_2\otimes\ldots\otimes X_d
\end{equation}
and the GP
kernel in \eqref{eq:gpkernel} is formed as a product of kernels across 
input dimensions 
\begin{equation}
K(X,X)=K_1(X_1)\otimes K_2(X_2)\otimes\ldots\otimes K_d(X_d).
\end{equation}
Under these conditions the storage requirements are reduced from $O(N^2)$ to $O(dN^{2/d})$ and the complexity of inversion is reduced from  $O(N^3)$ to $O(dN^{(d+1)/d})$, where $N$ is the cardinality of the full tensor
grid, i.e., the number of data points, and $N^{1/d}$ is the number of input points in each dimension. 
%The computationally expensive steps in GPR, e.g. estimating 
%the mean and variance in Eq.~\ref{eq:gpreg}, can be accelerated by exploiting 
%Kronecker structure, reducing the number of operations from $O(n^3)$ to 
%$O(Dn^{(D+1)/D})$. 

We review the key Kronecker algebra results, including efficient matrix-vector 
multiplication and eigendecomposition. For matrix-vector operations
\[
(A\otimes B)X=\mathrm{vec}\left(B\,X\,A^\top\right)
\]
where $v = \mathrm{vec}(V)$ converts column-major formatted matrices to vectors.
For higher dimensions, the expression above is applied recursively. In this 
approach, the full matrix is never formed, and individual steps rely only on
operations with individual kernels $K_i$. To compute the inverse 
$(K(X,X)+\sigma^2 I)^{-1}$ in \eqref{eq:gpreg}, we use the eigendecomposition
for each kernel  $K_i=Q_i^\top\Lambda_i Q_i$, which results in
\[
K+\sigma^2 I =(Q_1^\top\otimes Q_2^\top\otimes\ldots \otimes Q_d^\top)(\Lambda_1\otimes\Lambda_2\otimes\ldots \otimes \Lambda_d
+\sigma^2 I )(Q_1\otimes Q_2\otimes\ldots \otimes Q_d).
\]
The inverse is evaluated as
\[
(K+\sigma^2 I)^{-1}=(Q_1^\top\otimes Q_2^\top\otimes\ldots \otimes Q_d^\top)
(\Lambda_1\otimes\Lambda_2\otimes\ldots \otimes \Lambda_d +\sigma^2 I )^{-1}(Q_1\otimes Q_2\otimes \ldots \otimes Q_d).
\]
In this framework, the inverse of the full matrix now consists of eigendecompositions
of smaller matrices.
%, greatly reducing the computational cost of GP regression.
% The more
% relevant aspect of this paper is the separation of the covariance matrix into
% a Kronecker structure, so that the covariance matrix is the Kronkecker product
% of covariance matrices in each dimension.  For this approach, the build points
% are grid points, which may be an attractive aspect for many of our simulations.

% Cosmin:  please augment, include your example results, comments 
% about placement of grid points (can we use hierarchical grids or do they 
% need to be full tensor product grids), etc. 

\section{Conclusion}
\label{sec:conclusion}
Interest in machine learning for scientific applications has intensified in recent 
years, in part due to advances in algorithms, data storage, and computational 
analysis capabilities \citep{baker2019workshop,stevens2020ai}.  Fundamental challenges still remain when developing 
and using a machine learning model for scientific applications which must satisfy physical 
principles.  Enforcing such principles as constraints helps ensure the behavior of the 
model is consistent with prior physical knowledge when queried in an extrapolatory 
region. In other words, in addition to supplementing limited or expensive scientific data, 
constraints help improve the generalizability of the model in ways that simply increasing dataset size may not. 
Many approaches have been developed to perform ``physics-informed machine learning.'' 
In this survey, we have focused on constraint implementation in Gaussian processes, 
which are popular as a machine-learned metamodels or emulators for a computational simulation. 

Our survey focused on several important classes of constraints for Gaussian processes.
These included positivity or bound constraints on the Gaussian processes in Section \ref{sec:bound_constraints}.
When positivity constraints are applied to the derivatives of a Gaussian process, they lead to monotonicity and  convexity constraints as in Section \ref{sec:monotonicity_constraints}. This is a special example of regression with a linear transformation of a Gaussian process, which is the basis of Gaussian processes constrained by linear differential equations reviewed in Section \ref{sec:pde_constraints}. We discuss boundary value constrained Gaussian processes in Section \ref{sec:boundary_constraints}. 
Throughout, we see that constraints can be enforced in an implicit way through data that satisfies the constraint, by construction of a tailored sample space, by derivation of a constrained covariance kernel, or by modifying the output or likelihood of the Gaussian process. The constraints may be enforced in a ``global sense'', at a finite set of ``virtual'' or ``auxiliary'' points, or only in an approximate sense. We have pointed to these aspects as key features distinguishing the constraints in this survey. 

Constraints introduce new practical challenges into the GPR framework. These include: the analytical construction of sample spaces, transformations, or covariance kernels that inherently provide constraints; the sampling of truncated multivariate normals or intractable posterior distributions that arise when using non-Gaussian likelihoods; increased data and covariance matrix size when enforcing constraints with ``virtual'' data that leads to expanded ``four-block'' covariance; calculation of eigenvalues/eigenfunctions in bounded domains with complex geometry; the placement of virtual points or construction of spline grids in higher dimensions; and maximum likelihood training (optimization) of the hyperparameters of constrained Gaussian processes. Numerical issues are the focus of Section \ref{sec:computation_considerations}. In that section, we have also reviewed established numerical strategies for accelerating GPR. Some of these techniques have been applied to constrained Gaussian processes in the literature, while others have not. In general, the adaptation of computational strategies to constrained GPR is a relatively new field, and best practices have not yet been established. Moreover, while several codebases have been developed for constrained GPR, such \texttt{GPStuff} for non-Gaussian likelihoods \citep{vanhatalo2013gpstuff} and the \texttt{lineqGPR} package \citep{lopez2018lineqgpr} for the spline approach including constrained MLE, constraints have not made their way into the most widely used production codes for GPR. Furthering these computational aspects of constrained GPR remains a promising area for future work. 

The field of constrained Gaussian processes has made significant advances over the past decade, and we expect significant development to continue. The purpose of this survey, while non-exhaustive, has been to catalog and investigate some of the more common approaches, guide the practitioner to identify which strategies are most appropriate for his or her needs, and point out the new computational challenges of constrained Gaussian processes and how they can be approached.

\section*{Acknowledgements}
This work was completed with funding granted under Sandia's Laboratory Directed 
Research and Development program.
Sandia National Laboratories is a multi-mission laboratory managed and operated
by National Technology and Engineering Solutions of Sandia LLC, a wholly owned subsidiary of Honeywell International Inc. for the U.S. Department of Energy's 
National Nuclear Security Administration under contract {DE-NA0003525}. 
This paper describes objective technical results and analysis. Any subjective views or opinions that might be expressed in the paper do not necessarily represent the views of the U.S. Department of Energy or the United States Government. Report number: SAND2020-6086J.

\bibliographystyle{plainnat}
\bibliography{gpldrd} 
 
\end{document}